\documentclass{article}

\usepackage{arxiv}

\usepackage[utf8]{inputenc} 
\usepackage[T1]{fontenc}    
\usepackage{hyperref}       
\usepackage{url}            
\usepackage{booktabs}       
\usepackage{amsfonts}       
\usepackage{nicefrac}       
\usepackage{microtype}      
\usepackage{graphicx}
\usepackage{doi}

\usepackage{biblatex}
\usepackage{amssymb,amsmath}

\usepackage{mathtools}

\usepackage[ruled, vlined]{algorithm2e}

\graphicspath{{figures/}}
\usepackage{subcaption}
\usepackage{float}

\usepackage{amsthm}

\newtheorem{theorem}{Theorem}

\usepackage{subfiles} 

\usepackage{color}

\usepackage[ruled, vlined]{algorithm2e}

\usepackage{mathtools}

\newcommand{\algo}[1]{\textsc{#1}}

\title{Lifelong Learning in Multi-Armed Bandits}

\author{%
  Matthieu Jedor\\
  Centre Borelli, ENS Paris-Saclay \& Cdiscount\\
  \texttt{matthieu.jedor@ens-paris-saclay.fr} \\
   \And
   Jonathan Louëdec \\
  Cdiscount \\
   \texttt{jonathan.louedec@cdiscount.com} \\
   \And
   Vianney Perchet \\
   CREST, ENSAE Paris \& Criteo AI Lab \\
   \texttt{vianney.perchet@normalesup.org} \\
}

\date{}

\renewcommand{\headeright}{}
\renewcommand{\undertitle}{}
\renewcommand{\shorttitle}{Lifelong Learning in Multi-Armed Bandits}

\begin{document}
\maketitle

\begin{abstract}
	Continuously learning and leveraging the knowledge accumulated from prior tasks in order to improve future performance is a long standing machine learning problem. In this paper, we study the problem in the multi-armed bandit framework with the objective to minimize the total regret incurred over a series of tasks. While most bandit algorithms are designed to have a low worst-case regret, we examine here the average regret over bandit instances drawn from some prior distribution which may change over time. We specifically focus on confidence interval tuning of UCB algorithms. We propose a bandit over bandit approach with greedy algorithms and we perform extensive experimental evaluations in both stationary and non-stationary environments. We further apply our solution to the mortal bandit problem, showing empirical improvement over previous work.
\end{abstract}


\section{Introduction}


The multi-armed bandit (MAB) is a simple instance of online learning problem with partial feedback (see \textcite{bubeck2012regret}; \textcite{lattimore2018bandit}; \textcite{slivkins2019introduction}, and references therein). 
In the stochastic bandit framework, a learning agent sequentially pulls arms and obtains noisy rewards. 
The goal of the agent is to maximize her expected cumulative reward, or equivalently, to minimize her regret, which is the difference between the cumulative reward of an oracle (that knows the mean rewards of arms) and the one of the agent.
There is thus a clear trade-off, called ``exploration vs exploitation'', that arises between gathering information on uncertain arms by pulling them (exploration) and leveraging the information obtained so far by greedily pulling the arm with the highest estimated reward (exploitation).
MAB has received a large attention recently due to its wide applications range and the theoretical guarantees associated with learning algorithms (see also \textcite{bouneffouf2019survey}, and references therein for a survey on practical applications of MAB).

Most of the work in the stochastic bandit literature focuses on developing algorithms with optimal regret, either asymptotically in the instance-dependent sense or in the worst-case, on some problem class (typically on bounded or subgaussian rewards).
While the theory guarantees that these algorithms will have a sublinear regret in all instances within the problem class, they will be overly conservative for the majority of these instances, leading to a large regret.
Additionally, on these ``easier'' instances, sophisticated algorithms are outperformed by simpler ones \cite{vermorel2005multi, kuleshov2014algorithms}, which is an obstacle to their implementation in practice. 
In this paper, we consider that the learner successively interacts with a task sampled from a problem class with some prior distribution, which may or may not be stationary over time.
Within each task, there is a learning problem which is a MAB problem with a fixed time horizon. 
The learning agent does not know the model parameters of each bandit problem, nor does she know the prior distribution within the problem class.
Yet, it is critical for the learner be to able to ``track'' this probability distribution to build lifelong learning agent and thus achieve the best possible performance.
In a stationary environment, the learner experiences the same bandit problem over and over again, and an efficient algorithm, in order to improve the performance when it faces the same problem again, should leverage the information acquired from previous tasks.
Conversely in a non-stationary environment, the distribution within the problem class may change over time and the previously learned solution may fail to keep its good performance under the new distribution. An efficient algorithm must thus continuously learn to improve itself.
The goal of the learning agent is then to minimize the lifelong regret, that is the cumulative regret across all tasks. 
This approach can be regarded as a special case of meta-learning \cite{schmidhuber1987evolutionary, thrun2012learning} or lifelong learning \cite{thrun1998lifelong, chen2016lifelong, silver2013lifelong}.

\paragraph{Our results} 
We focus on a tractable instance of this problem that is the tuning of the confidence interval width of \algo{UCB}-like policies. We first evaluate the impact of the choice of algorithm on empirical performance over several problem classes with different prior and different number of arms. On a side note, we prove a bound on the Bayesian regret of a tuned \algo{UCB} algorithm showing that we do not lose all theoretical guarantees. 
We then concentrate on learning in a stationary environment. We first investigate the influence of various initializations on performance. Next, we discuss about the main part of this paper that is learning the optimal algorithm. For this purpose, we consider a bandit over bandit approach, that is using a bandit algorithm to choose the optimal parameter. We show empirically that the \algo{Greedy} algorithm as the meta-algorithm performs extremely well. 
We next concentrate on learning in a non-stationary environment, precisely we look at both an abruptly changing and a slowly changing environment. To this effect, we adapt the \algo{Greedy} algorithm using methods from non-stationary bandits.
Finally, we apply our method to a more realistic setting, the mortal bandit problem. By decomposing the time horizon into episodes according to the expected lifetime of arms, we show great empirical improvement compared to previous work.


\paragraph{Related Work}


Bayesian bandits \cite{gittins1979bandit, gittins2011multi, berry1985bandit} have been studied extensively with the goal of developing optimal algorithms in the Bayesian sense.
Lower bounds on the Bayesian regret have also been derived \cite{lai1987adaptive, kaufmann2016bayesian}.
Unfortunately, computing the Bayesian optimal algorithm is generally intractable \cite{lattimore2018bandit}.

The closest to our work is that of \textcite{lazaric2013sequential} who considered a framework which closely resembles to ours except they studied the stochastic setting and further assumed a finite set of models; whereas we consider the Bayesian setting and do not make any assumption on the number of changes in the prior distribution. 
\textcite{deshmukh2017multi} extended the previous work in the contextual framework. 

Also related to our work, \textcite{maes2012meta} tuned existing algorithms and also learned index algorithms of historical features. 
\textcite{hsu2019empirical} proposed a best-arm identification algorithm for tuning the confidence interval of the \algo{UCB} algorithm and the posterior distribution of the \algo{Thompson Sampling} algorithm. 
\textcite{boutilier2020differentiable} focused on ``differentiable'' algorithms and optimized them by gradient ascent.
In comparison, the setting of these works are offline while we are more interested in the lifelong regret incurred in potentially non-stationary environments.

Several works also  use bandit algorithms as meta-algorithm.
\textcite{li2017hyperband} introduced a bandit-based approach to hyperparameter optimization using \algo{Sequential Halving}, a pure-exploration bandit algorithm, as a subroutine.
The celebrated \algo{UCT} algorithm \cite{kocsis2006bandit} makes use of the \algo{UCB} algorithm applies on trees. 
In the non-stationary framework, a few methods involve MAB in a hierarchical way: \textcite{hartland2006multi} considered a meta-bandit to decide whether to accept the change detection or not; \textcite{cheung2019hedging} used the \algo{EXP3} algorithm to decide the window size of the \algo{SW-UCB} algorithm; and \textcite{wu2018learning, wu2019dynamic} adopted a hierarchical bandit algorithm where a master bandit manages some slave bandits.

The model can also be viewed as a special case of mortal bandits \cite{chakrabarti2009mortal, bnaya2013volatile, traca2019reducing} in which arms show up by batch where the time of death is the same for each arm in a given batch.



\section{Setting}


In the stochastic MAB model, an agent interacts sequentially with a set of $K$ unknown distributions $\mathcal{V}_1, \dots, \mathcal{V}_K$, called arms. At time $t$, the agent chooses an arm $A_t$ which yields a reward $X_t$ drawn from the associated probability distribution $\mathcal{V}_{A_t}$. 
The objective is to design a sequential algorithm maximizing the expected cumulative reward up to some time horizon $T$. Let $\mu_1, \dots, \mu_K$ denote the mean rewards of arms, and $\mu^\star \coloneqq \max_{k \in [K]} \mu_k$. The goal is equivalent to minimizing the regret, defined as the difference between the expected reward accumulated by an oracle strategy always playing the best arm at each round, and the one accumulated by an algorithm $\mathcal{A}$,
\begin{equation*}
    R(T, \mathcal{A}) = \mathbb{E} \left[ \sum_{t=1}^T \left( \mu^\star - \mu_{A_t} \right) \right]
\end{equation*}
where the expectation is taken with respect to the randomness in the sequence of successive rewards from each arm and the possible randomization of the algorithm.
Furthermore, we assume the following Bayesian, parametric bandit setting: a random vector $\boldsymbol{\theta} = \left( \theta_1, \dots, \theta_K \right)$ is drawn from a prior distribution $\Pi$, and the distribution of arm $k$ depends on the parameter $\theta_k$. This leads to the notion of Bayesian regret which depends on the prior distribution $\Pi$,
\begin{equation*}
    \mathrm{BR}_\Pi(T, \mathcal{A}) = \int R(T, \mathcal{A}) d\Pi(\boldsymbol{\theta}) \,.
\end{equation*}
We consider a lifelong learning setting where at each episode $j$ the learner interacts with a task $\mathcal{V}_{\theta_1^j}, \dots, \mathcal{V}_{\theta_K^j}$, where each $\theta_k^j$ is drawn i.i.d.\ from a prior distribution $\Pi_j$. The objective is to find a series of algorithms minimizing the lifelong (Bayesian) regret over $J$ episodes, 
\begin{equation*}
    \mathrm{LR}_{\boldsymbol{\Pi}}(T, \mathfrak{A}) = \sum_{j=1}^J \mathrm{BR}_{\Pi_j}(T, \mathcal{A}_j)
\end{equation*}
where $\mathfrak{A} = \left( \mathcal{A}_1, \dots, \mathcal{A}_J \right)$ and $\boldsymbol{\Pi} = \left( \Pi_1, \dots, \Pi_J \right)$ denote respectively the algorithms and the prior distributions for each episode. We further assume that the horizon for each task $T$ and the number of episodes $J$ are known. An unknown number of episodes can be handled as usual \cite{Degenne}; 
while the knowledge of the time horizon is mostly for convenience in the choice of the sub-algorithm and similar results can be achieved with an anytime algorithm.

In order to minimize the lifelong regret, our method
consists in designing a meta-algorithm 
which, at each episode $j$, selects an algorithm $\mathcal{A}_j$ from a class of algorithms, which can be finite or infinite, that aims at minimizing the Bayesian regret with respect to the prior distribution $\Pi_j$.

    
    

We especially focus on a tractable class of algorithms, the \algo{UCB} algorithm \cite{auer2002finite} with a parameter $\gamma$ controlling the width of the confidence interval. We emphasis that similar results can be obtained with any \algo{UCB}-like algorithms. Formally, the \algo{UCB}($\gamma$) index of arm $k \in [K]$ in round $t$ is 
\begin{equation*}
    U_t(k) = \widehat{\mu}_{k}(t-1) + \gamma \sqrt{\frac{2 \log\left( 1 / \delta\right)}{N_k(t-1)}}
\end{equation*}
where $\widehat{\mu}_{k}(t)$ is the average reward of arm $k$ in the first $t$ rounds, $N_k(t)$ is the number of times that arm $k$ has been pulled in the first $t$ rounds and $\delta = 1/T$ is the probability that the confidence interval fails (in subsequent sections, we will choose $\delta = 1/T$). 
Note that the case $\gamma = 1$ corresponds to the theoretical value that attains the optimal instance-dependent regret for Gaussian rewards, while $\gamma = 0$ corresponds to the \algo{Greedy} algorithm. 
Hence for this class of algorithms, the meta-algorithm faces a new trade-off between being conservative and being aggressive on a class of bandit instances.
Thereby, the objective that is to minimize the lifelong regret is equivalent to finding the optimal values of $\gamma \in [0, 1]$ that minimizes the Bayesian regret in each episode.

Theoretically, any worst-case optimal algorithm will also be optimal, up to constant factors, in the Bayesian setting \cite{lattimore2018bandit}; and thus optimal in the lifelong setting. Yet, we will see that we can empirically improve these algorithms.
We end this section with an analysis of this tuned \algo{UCB}.
More generally, \textcite{russo2014learning} noticed that the Bayesian regret of any \algo{UCB}-like algorithm satisfies
\begin{equation}
    \mathrm{BR}_\Pi(T, \algo{UCB}) \leq \mathbb{E}\left[ \sum_{t=1}^T \left( \mu^\star - U_t(A^\star) \right) + \sum_{t=1}^T \left(U_t(A_t) - \mu_{A_t} \right) \right]
    \label{eq:decomp_bayes_regret}
\end{equation}
where $A^\star$ denotes the optimal arm, which is a random variable. Specifically for the \algo{UCB}($\gamma$) algorithm we state the following theorem.

\begin{theorem}
For 1-subgaussian distributions with mean in $[0,1]$, \algo{UCB}($\gamma$) with $\gamma > 0$ satisfies
\begin{equation*}
    \mathrm{BR}_\Pi(T, \algo{UCB}(\gamma)) \leq 4 K T^2 \delta^{4 \gamma^2} + 2 \gamma \sqrt{2KT\log(1/\delta)} \,. 
\end{equation*}
\end{theorem}

As an example, for $\gamma = 1$ and $\delta = 1 / T$, we get the known bound $\mathcal{O}\left( \sqrt{KT\log{}T} \right)$.

\begin{proof}
Let $E$ be the event that for all $t = 1, \ldots, T$ and $k = 1, \ldots, K$,
\begin{equation*}
    \left| \widehat{\mu}_k(t-1) - \mu_k \right| < \gamma \sqrt{\frac{2 \log(1/\delta)}{N_k(t-1)}} \,.
\end{equation*}
Using the subgaussian assumption and a union bound, we get that $\mathbb{P}\left( E^c \right) \leq 2KT\delta^{4\gamma^2}$. On the event $E^c$, the terms inside the expectation of Equation \eqref{eq:decomp_bayes_regret} are bounded by $2T$, while on the event $E$, the first sum is bounded by 0 and for the second term, standard computations yield
\begin{equation*}
    \sum_{t=1}^T \left(U_t(A_t) - \mu_{A_t} \right) \mathbf{1}\{E\} \leq 2 \gamma \sqrt{2KT\log(1/\delta)} \,.
\end{equation*}
Putting together the pieces completes the proof.
\end{proof}


\section{Choice of the class of algorithms} \label{sec:choice_strategies}

\begin{figure}[t]
    \begin{subfigure}{0.33\textwidth}
    \includegraphics[width=\linewidth]{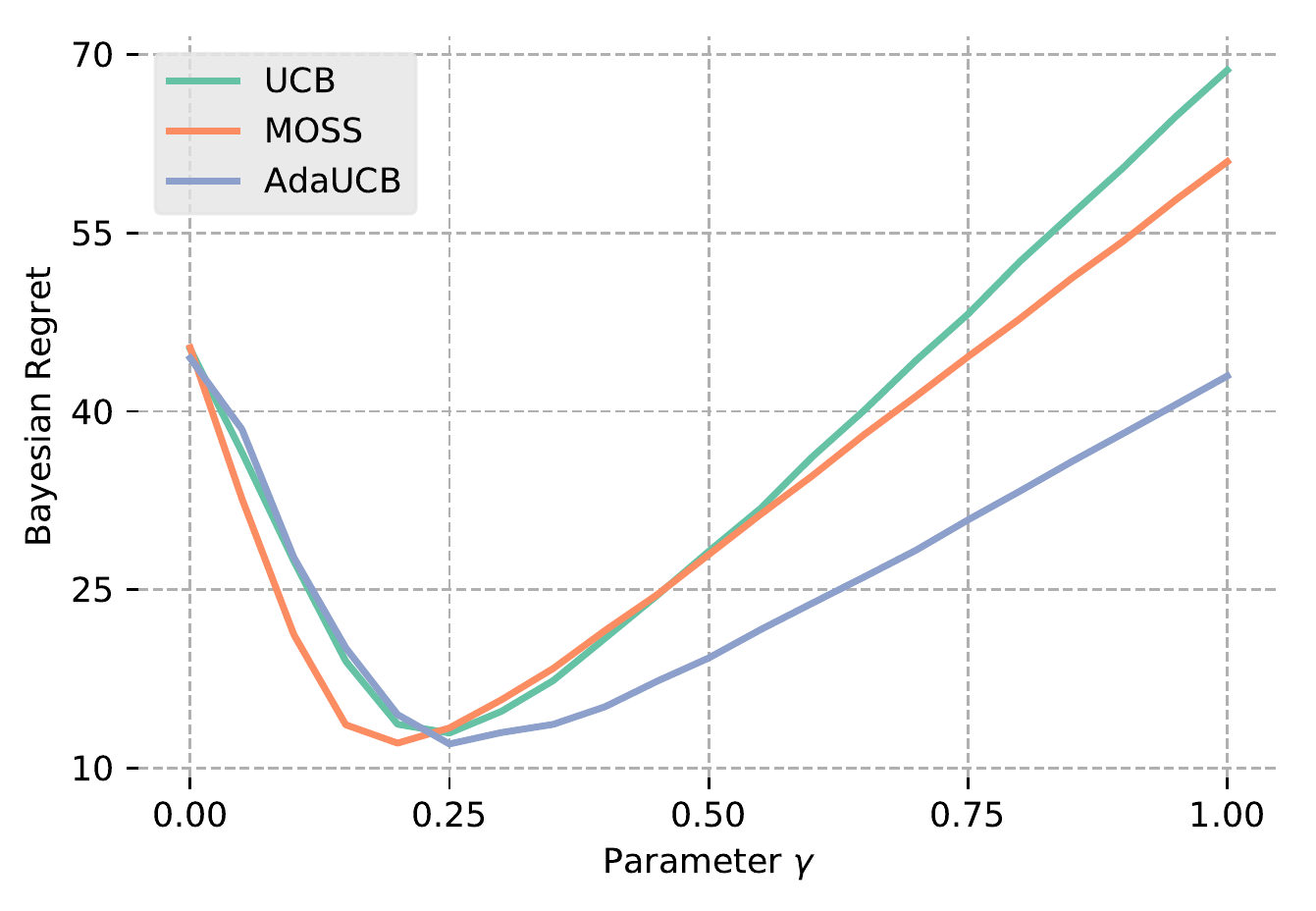}
    \caption{$K=5$}
    \label{fig:choice_sub_policy_a}
    \end{subfigure}
    \begin{subfigure}{0.33\textwidth}
    \includegraphics[width=\linewidth]{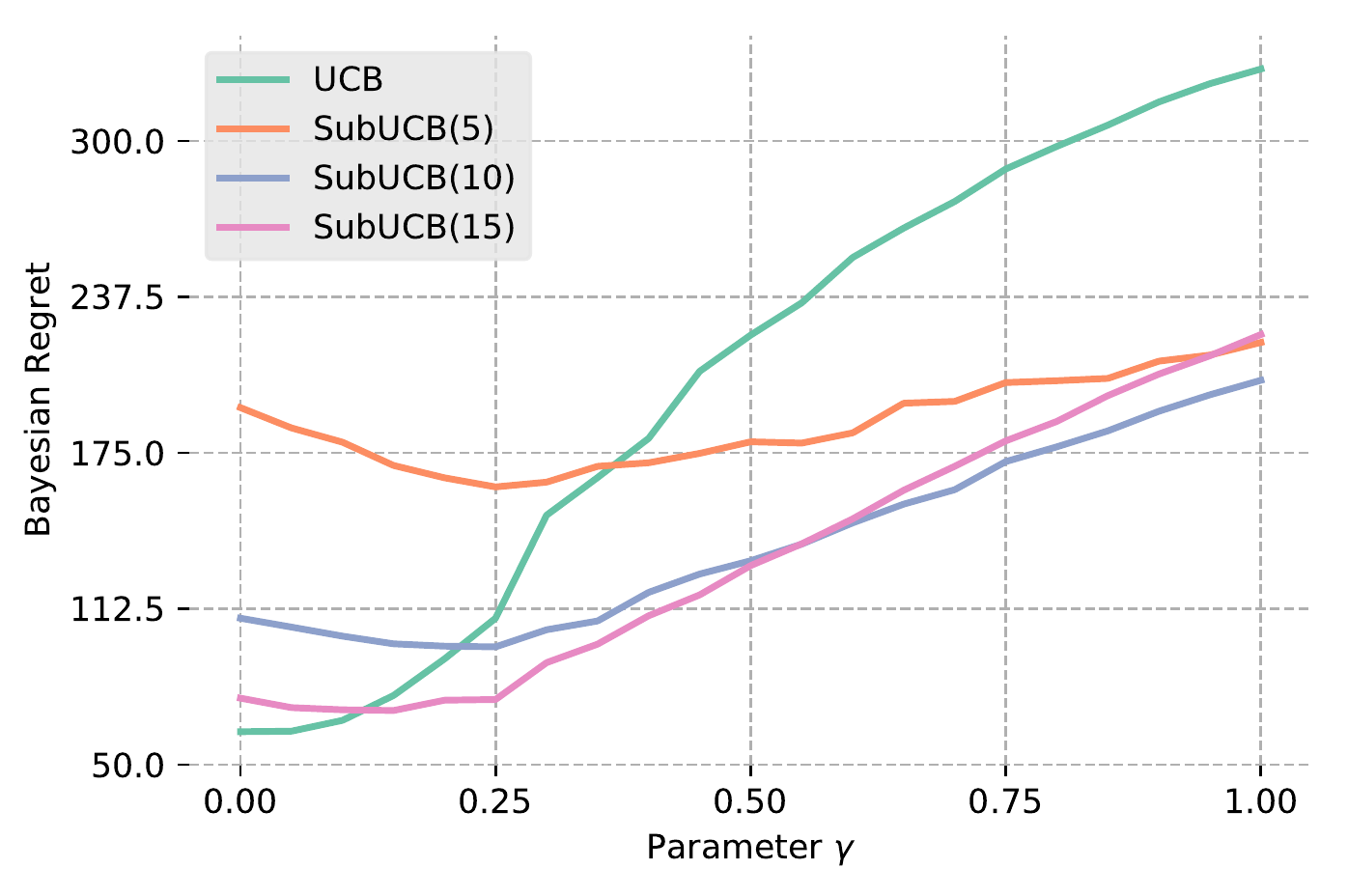}
    \caption{$K=63$}
    \label{fig:choice_sub_policy_b}
    \end{subfigure}
    \begin{subfigure}{0.33\textwidth}
    \includegraphics[width=\linewidth]{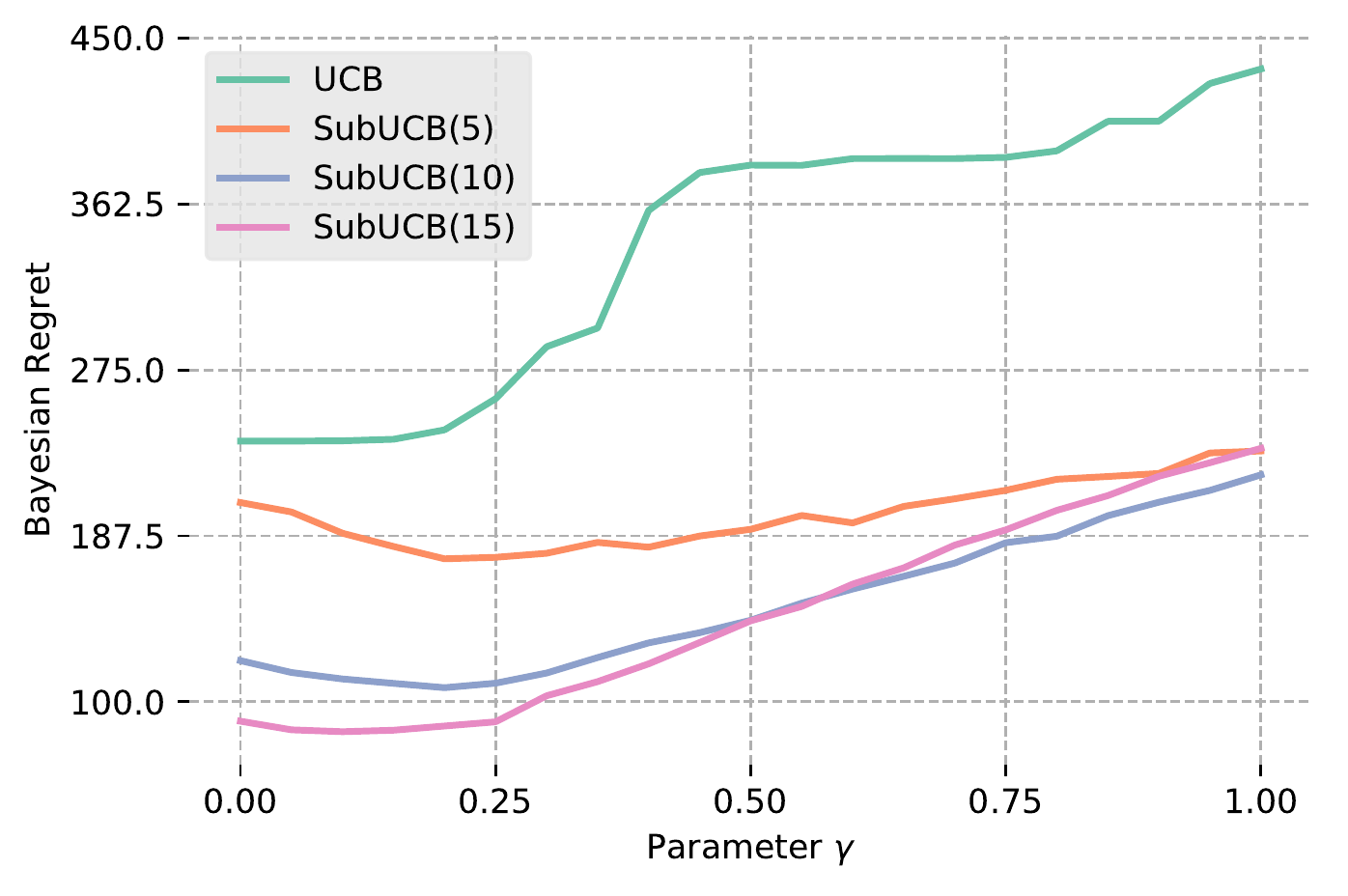}
    \caption{$K=250$}
    \label{fig:choice_sub_policy_c}
    \end{subfigure}
    
    \caption{Bayesian regret of various algorithms as a function of $\gamma$ for diverse numbers of arms $K$.}
    \label{fig:choice_sub_policy}
\end{figure}

We begin by showing that the choice of the class of algorithms in which we seek to find the one that maximize the average cumulative reward over bandit instances, is of the utmost importance in order to have the best possible empirical performance, and the most favorable algorithm depends on a number of factors. 
In Figure \ref{fig:choice_sub_policy}, we evaluate the Bayesian regret of several \algo{UCB}-like algorithms as a function of their parameter $\gamma$ in different setups. In all experiments, the distribution of each arm is a Bernoulli distribution where the mean reward is drawn i.i.d.\ from a uniform distribution over $[0, 1]$. The horizon is fixed at $T=1000$ and we vary the number of arms $K$. Results are averaged over $5000$ iterations. See also Appendix \ref{appendix:choice_class_strategies} for additional experiments with different distributions and priors.

In Figure \ref{fig:choice_sub_policy_a}, we compare, for a small number of arms $K$, different \algo{UCB}-like algorithms: the original UCB algorithm \cite{auer2002finite}, \algo{MOSS} \cite{audibert2009minimax}, which is known to be worst-case optimal contrary to \algo{UCB}, and \algo{AdaUCB} \cite{lattimore2018refining} which is simultaneously worst-case optimal, asymptotically optimal, and never worse than \algo{UCB} in the worst case. Interestingly, the later statement is also verified on the Bayesian regret for all values of $\gamma$. The same can also be said almost everywhere for \algo{MOSS} over \algo{UCB}. For large values of $\gamma$, \algo{AdaUCB} outperforms \algo{MOSS} while for small values, \algo{MOSS} only improves slightly over \algo{AdaUCB}.
As a result, if one is purely interested in great empirical performance, a default choice would be \algo{AdaUCB} for any value of $\gamma$. 
However in the lifelong setting, we want to optimize the parameter $\gamma$ and it is not clear which choice of algorithm will perform better through all the episodes; \algo{AdaUCB} may be better but this also makes it harder for the meta-algorithm to find the optimal value of $\gamma$, meaning a potentially larger lifelong regret compared to \algo{UCB} for example.
Additionally, the best Bayesian regret of the optimally tuned version of each algorithm is roughly the same for all three.
Hence our choice of the \algo{UCB} algorithm, which is simple enough for illustration, yet may also be suitable for practical applications.
We finally note that for all algorithms, the gain of their tuned version over the default choice ($\gamma = 1$) is substantial even though \algo{MOSS} and \algo{AdaUCB} are worst-case optimal up to constant factors, supporting our choice to optimize algorithms for practical purpose.

In Figure \ref{fig:choice_sub_policy_b} and \ref{fig:choice_sub_policy_c}, we compare the \algo{UCB} algorithm with sub-sampled versions of itself, denoted \algo{SubUCB}, in the case of an intermediate and a large number of arms $K$. Formally, \algo{SubUCB}($m$) selects randomly $m$ arms and performs \algo{UCB} on these arms. For $K=63$, we see that the \algo{Greedy} algorithm, symbolized by UCB with $\gamma = 0$, performs better than any tuned \algo{UCB} and also better than \algo{SubUCB}. For $K=250$, \algo{SubUCB} performs always better than \algo{UCB}. It is interesting to notice that as $m$ grows larger, up to a certain point, \algo{SubUCB} has a better optimal Bayesian regret, becomes more sensitive to the parameter $\gamma$ and a sub-sampled version of the \algo{Greedy} algorithm turns into the best performing algorithm.
We mention that these behaviors, i.e.\ the good empirical performance of the \algo{Greedy} algorithm and the superiority of a sub-sampled version of \algo{UCB} in the case of a great number of arms, have been pointed out by \textcite{bayati2020optimal}. We showed that they also hold true empirically after tuning of the confidence interval width.


\section{Learning in a stationary environment}


In this section, we consider the learning of the optimal algorithm in a fixed environment, i.e.\ when the prior distribution is the same in all episodes. In this case, the objective is equivalent to finding the algorithm that minimizes the Bayesian regret for that specific prior distribution.

\subsection{Influence of the initialization} \label{sec:fixed_init}

We start with a study of the effect of the initialization choice on empirical performance.
By default, most bandit algorithms initialize arms by pulling them at least one time. Knowing that we solve again and again similar bandit problems, we may want to find a more clever initialization. For example, consider a challenging bandit problem with a large number of arms with respect to the time horizon and assume we have found a reasonably good arm; exploiting this arm may then be more rewarding that exploring new arms in the hope of finding a better one. This becomes especially critical the more greedy we get.
In this section, we fix the value of the hyperparameter $\gamma$ and we evaluate three different initializations. In all cases, we set the empirical means of arms to a specific value and their upper confidence bounds are built as if they have been played once. In the first case we set this value at 0 (Init 1), in the second (Init 2) and third (Init 3) cases, it is fixed at the mean and median, respectively, of previous empirical means. We denote by ``Init 0'' the default initialization.

\begin{figure}[t]
    \begin{subfigure}{0.49\textwidth}
    \includegraphics[width=\linewidth]{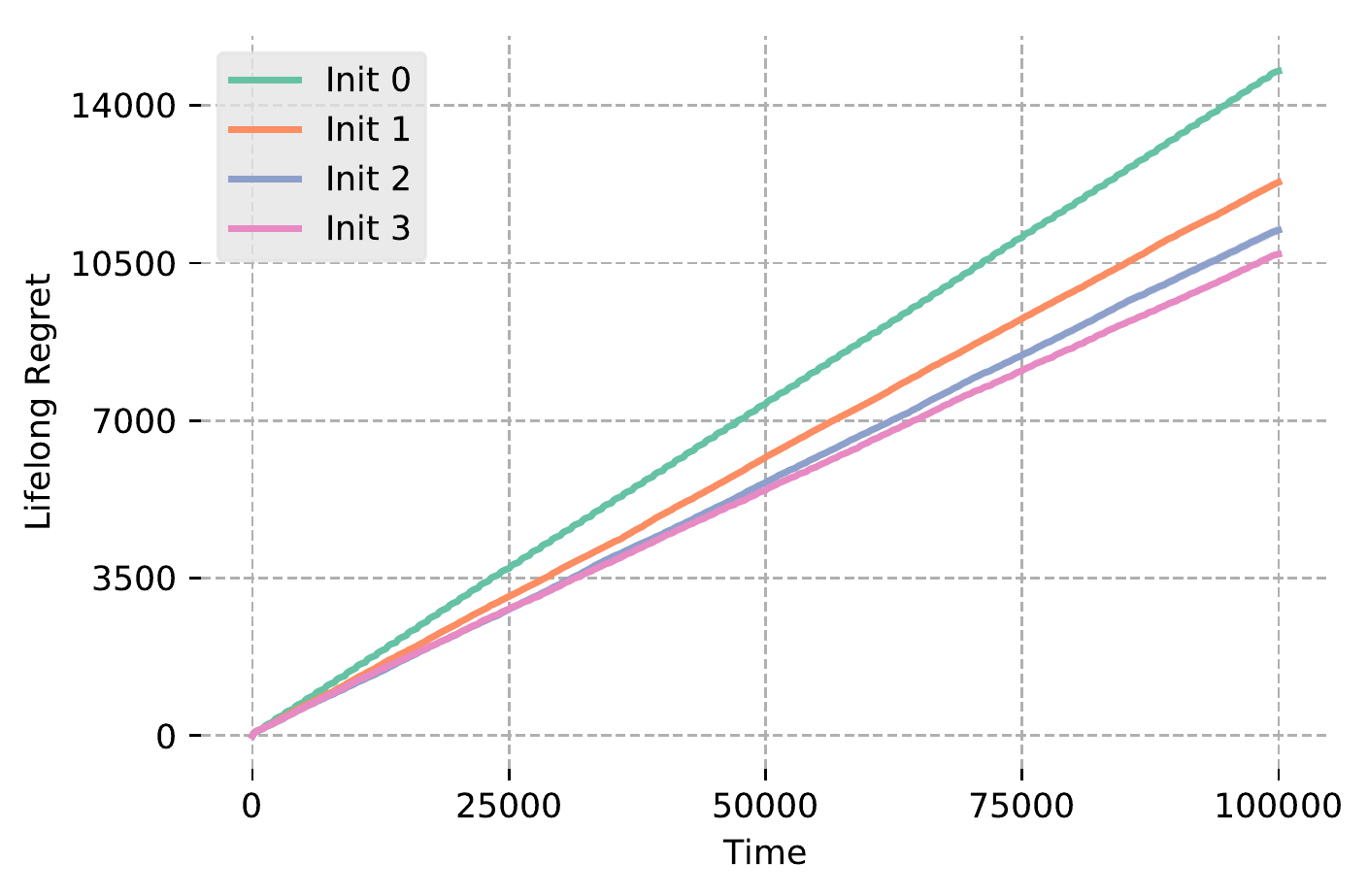}
    \caption{Comparison of various initializations}
    \label{fig:fixed_env_init}
    \end{subfigure}
    \begin{subfigure}{0.49\textwidth}
    \includegraphics[width=\linewidth]{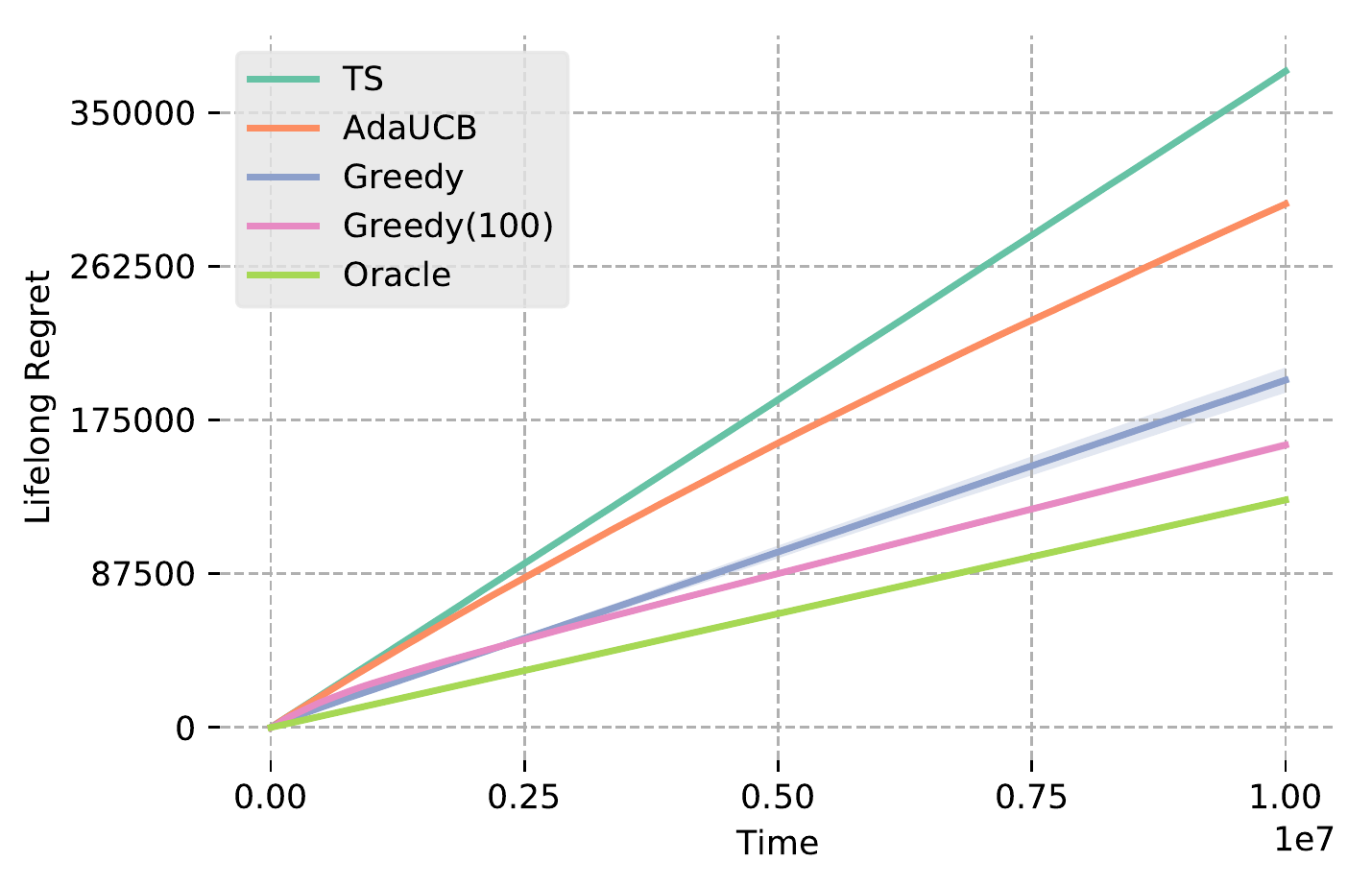}
    \caption{Comparison of various meta-algorithms}
    \label{fig:fixed_env_learning}
    \end{subfigure}
    \caption{Lifelong regret of various meta-algorithms in stationary environments. Shaded areas show standard errors.}
\end{figure}

We analyze here a particular problem class and we refer the reader to Appendix \ref{appendix:init} for a more complete overview on the impact of the different initializations for different prior distributions and number of arms. In this experiment, each episode consists of a Gaussian bandit problem with $K=5$ arms, a time horizon $T=,000$ and the mean rewards of arms are drawn i.i.d.\ from a uniform distribution over $[0, 1]$. We fix $\gamma = 0.2$ and we repeat this tuned UCB for $J=100$ episodes with the different initializations previously mentioned. On Figure \ref{fig:fixed_env_init}, we report the lifelong regret averaged over 100 iterations.

In this experiment, all three initializations improve over the default choice; it turns out that initializing arms with the median of previous empirical means performs better than the mean of previous arms which in turn performs better than an initialization at 0. However, this is not always the case and it heavily depends on the number of arms $K$ and on the prior distribution. Fortunately, this choice is insignificant on instances with a small number of arms, which we study in this paper, therefore in what follows we assume that \algo{UCB} uses the default initialization in each episode.

\subsection{Bandit algorithms as meta-algorithm} \label{sec:fixed_learning}

Now that the initialization rule is set, we can focus on the meta-algorithm, i.e.\ the algorithm that is responsible for picking the parameter $\gamma$ of \algo{UCB} for each episode, and ultimately, is the keystone in the minimization of the lifelong regret. We consider bandit algorithms for the choice of the meta-algorithm, since they are efficient online optimization algorithms.
This may seem like a vicious circle as we are talking about optimization of bandits algorithms and we want to avoid having to optimize the optimizer. Fortunately, the two algorithms, the meta-algorithm and the sub-algorithm, face a different problem. Indeed, the sub-algorithm aims at maximizing the average reward over bandit instances while the meta-algorithm aims at maximizing a function, which is the expected cumulative reward of the sub-algorithm with respect to its parameter $\gamma$.
The dilemma encountered by the meta-algorithm is actually a continuous-armed bandit problem \cite{kleinberg2005nearly, auer2007improved} where the set of arms lies in some bounded interval, in our case the different $\gamma \in [0, 1]$. \textcite{kleinberg2005nearly} proposed a simple, yet nearly optimal, algorithm which consists in discretizing the $[0, 1]$ interval into a finite set of $n$ equally spaced points and running a standard bandit algorithm over those points. Unfortunately, their theoretical result holds only when the function to be optimize satisfies some Hölder conditions which may not verified for the Bayesian regret of \algo{UCB}($\gamma$). Still, that does not refrain us from using this strategy. The chosen number of arms $n$ is critical in practice; set too low we may be far from the optimal solution and set too high we may end up exploring all the time. \textcite{auer2007improved}, with a similar algorithm, claimed a value $n = (J / \log J)^{1/3}$ is optimal without knowing the exact Hölder condition. We thus choose this specific discretization in our simulations. It has also been noted by \textcite{bayati2020optimal} that the \algo{Greedy} algorithm, known to have a linear regret, ran with a sufficiently large number of arms may benefit from ``free'' exploration. This change point in its behavior happens around $\sqrt{J}$; we also evaluate \algo{Greedy} with a discretization which contains that many points.

We again study a particular bandit instance; see Appendix \ref{appendix:fixed_env} for the same experiment made with different choices of prior distribution. In this experiment, each episode consists of a Bernoulli bandit problem with $K=5$ arms, a time horizon $T=1000$ and the mean rewards are drawn i.i.d.\ from a uniform distribution over $[0, 1]$. We set the number of episodes $J=10000$ and we compare different meta-algorithms, namely \algo{Thompson Sampling} (\algo{TS}) with a uniform prior \cite{agrawal2013further}, \algo{AdaUCB} \cite{lattimore2018refining} and the \algo{Greedy} algorithm with the two discussed discretizations, denoted \algo{Greedy}(100) for the discretization with 100 points. We also report an oracle meta-algorithm, which knows the optimal $\gamma$. Results are averaged over 100 iterations and are displayed on Figure \ref{fig:fixed_env_learning}.

We see that \algo{TS} has a ``linear'' regret indicating that it fails to learn a good parameter $\gamma$ within that time frame. Whereas the regret of \algo{AdaUCB} is sublinear and the algorithm is thus learning; however it is outperformed for a relatively long period of time by a naive \algo{Greedy}, which is stuck to a suboptimal, yet good arm. The most interesting part is that \algo{Greedy}(100) performs extremely well; its regret is remarkably close to the one of \algo{Oracle} and is even sublinear. This supports the notion of free exploration for a large enough number of arms of the \algo{Greedy} algorithm.


\section{Leaning in a non-stationary environment}


We now focus our attention to learning in a non-stationary environment, i.e.\ in which the prior distribution is not the same for all episodes. 
We will consider two scenarios: one where the environment changes abruptly and another where it slowly changes over time.
Regrettably, adaptively tracking the prior distribution is intractable without making strong distributional assumptions (and potentially at the cost of forced exploration). Nonetheless, that does not mean that trying to improve over a naive meta-algorithm is hopeless. Taking inspiration from the stochastic non-stationary bandit literature \cite{garivier2011upper}, we adapt the index of \algo{Greedy} to take into account a changing environment.

The following two adjustments are based on the idea of ``forgetting'' rewards obtained long ago. The first one is based on discounting. Formally, let $\omega \in (0,1)$ be the discount factor and define the discounted \algo{DGreedy} index
\begin{equation*}
    \widehat{\mu}_k^\omega(t) = \frac{\sum_{s=1}^t \omega^{t-s} X_k \mathbf{1} \{ A_s = k \}}{\sum_{s=1}^t \omega^{t-s} \mathbf{1} \{ A_s = k \}} \,.
\end{equation*}
The idea is to reduce the weight of rewards collected a long time ago, making the index more sensitive to recent rewards. A similar approach, somewhat more sharp, called sliding-window, simply discards rewards older than a parameter $\tau \in \mathbb{N}^\star$. Formally, the \algo{SWGreedy} index is defined as
\begin{equation*}
    \widehat{\mu}_k^\tau(t) = \frac{\sum_{s = t - \tau + 1}^t X_k \mathbf{1} \{ A_s = k \}}{\sum_{s = t - \tau + 1}^t  \mathbf{1} \{ A_s = k \}} \,.
\end{equation*}
Both these indexes are similar to those of the \algo{Discounted UCB} and \algo{Sliding-Window UCB} \cite{garivier2011upper}, the difference being that the bonus terms are removed. 
Unfortunately, there is no choice of $\omega$ and $\tau$ which guarantees strong performance, and they must be tuned empirically for the specific environment. Additionally, because both of this algorithms do not reflect on the whole horizon, the optimal discretization for the \algo{Greedy} meta-algorithm is also more challenging to determinate. Consequently, in the following, we set the number of points in the discretization at a lesser value.

\begin{figure}[t]
    \begin{subfigure}{0.49\textwidth}
    \includegraphics[width=\linewidth]{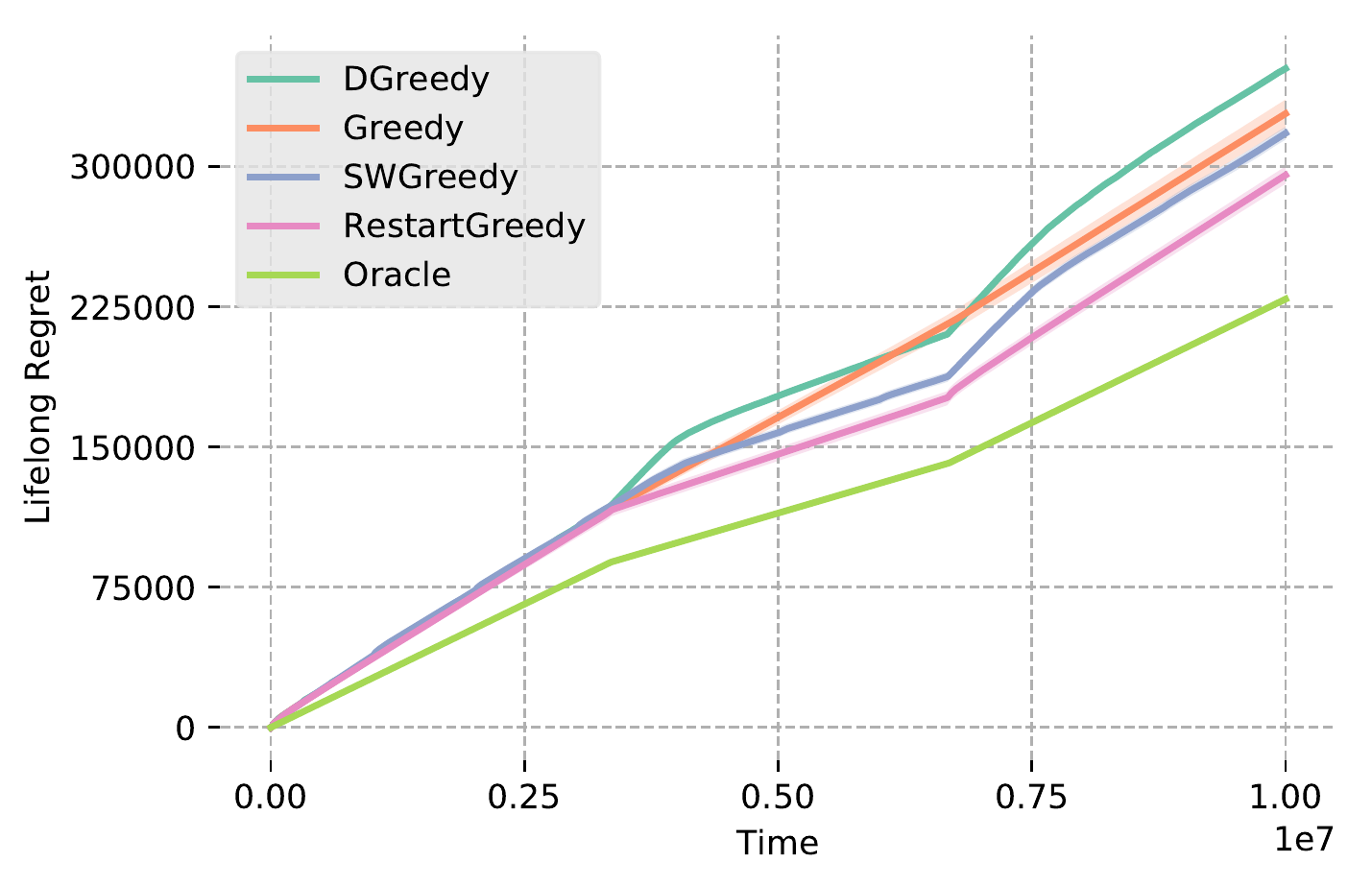}
    \caption{Abruptly changing environment}
    \end{subfigure}
    \begin{subfigure}{0.49\textwidth}
    \includegraphics[width=\linewidth]{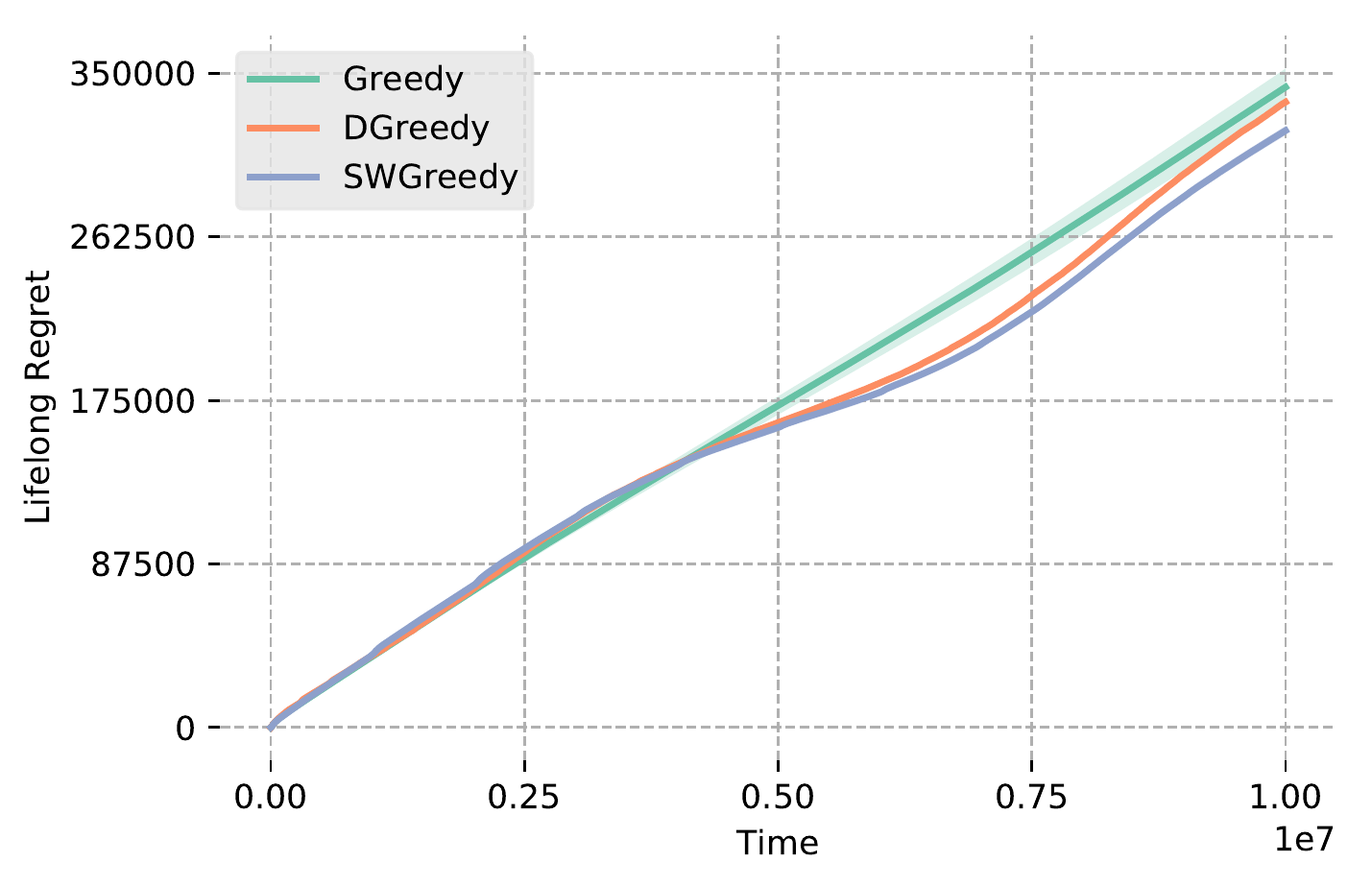}
    \caption{Slowly changing environment}
    \end{subfigure}
    \caption{Lifelong regret of various meta-algorithms in non-stationary environments. Shaded areas show standard errors.}
    \label{fig:changing_env}
\end{figure}

In the following experiments, we run the \algo{Greedy} meta-algorithm with the different indexes\footnote{We choose $\omega=0.9975$ and $\tau = 1,000$ on both scenarios; these parameters have been roughly tuned.} on a discretization consisting of 21 equally-spaced points. In both scenarios, each episode is a Bernoulli bandit problem with $K=10$ arms, the time horizon $T=1000$ and the number of episodes is set at $J=10000$.
In the first scenario, we consider an abruptly changing environment, the prior distribution over the mean rewards of arms is a Beta(1, 3) distribution for $j \leq J/3$ and $j \geq 2J/3$ and a Beta(3, 1) distribution in between. While in the second scenario, we consider a slowly changing environment, the prior distribution at episode $j$ is a Beta($2+\cos(2\pi j/J+\pi), 2+\cos(2\pi j/J)$). In the first scenario, we also report an algorithm which restart at times where the prior changes; this can be seen as an oracle we aim to emulate. Results are reported on Figure \ref{fig:changing_env} and are averaged over 100 runs.

In both scenarios, \algo{DGreedy} and \algo{SWGreedy} are able to track the change in the prior distribution while the naive \algo{Greedy} fails to do so. 
In the abruptly changing scenario, \algo{SWGreedy} outperforms \algo{DGreedy}, thus underlining the need to discard data from the previous prior. Regrettably, when the first prior comes back, both algorithms take a long time before finding again the previous learned solution; making the naive \algo{Greedy} competitive in this experiment even if it completely missed the first change.
Conversely in the slowly changing scenario, \algo{Greedy} fails to track the slow change in the prior distribution and thus accumulates a larger regret with in addition a broad variance. \algo{DGreedy} and \algo{SWGreedy} are roughly similar in this case, though \algo{SWGreedy} performs slightly better in the end.
We would also like to mention that there is a hidden parameter in \algo{DGreedy}. Indeed, in the \algo{Greedy} algorithm, each arm must have enough information for its index to calculated. In \algo{Greedy}, and also \algo{SWGreedy}, this is done by ensuring that each arm has been played at least once. In \algo{DGreedy}, this initialization is more challenging to determine and we believe it is partly responsible for the poor performance of \algo{DGreedy}. In both experiments, we arbitrarily set this value to 1.


\section{Application to mortal bandits}


Finally in this last section, we apply our method to a more realistic setting, the mortal bandit problem \cite{chakrabarti2009mortal}, where arms appear and disappear regularly (in particular, an arm is not always available contrary to the standard model). 
While the notion of episode may be more elusive in this setting, we show that synchronizing an episode according to the expected lifetime of arms can help to overcome this difficulty. We look upon the case where this  value is known and the one where it is not and thus have to be estimated. We also have to modify the index of the \algo{UCB} sub-algorithm to take into account that arms arrive at different times and for the unknown expected lifetime. Thereby, we replace the $\log(T)$ term by $\log(t - s_k + 1)$, where $t$ and $s_k$ denotes respectively the current and the arrival time steps for arm $k$. We then again pick the \algo{Greedy} algorithm as meta-algorithm.

\begin{figure}[t]
    \begin{subfigure}{0.49\textwidth}
    \includegraphics[width=0.99\linewidth]{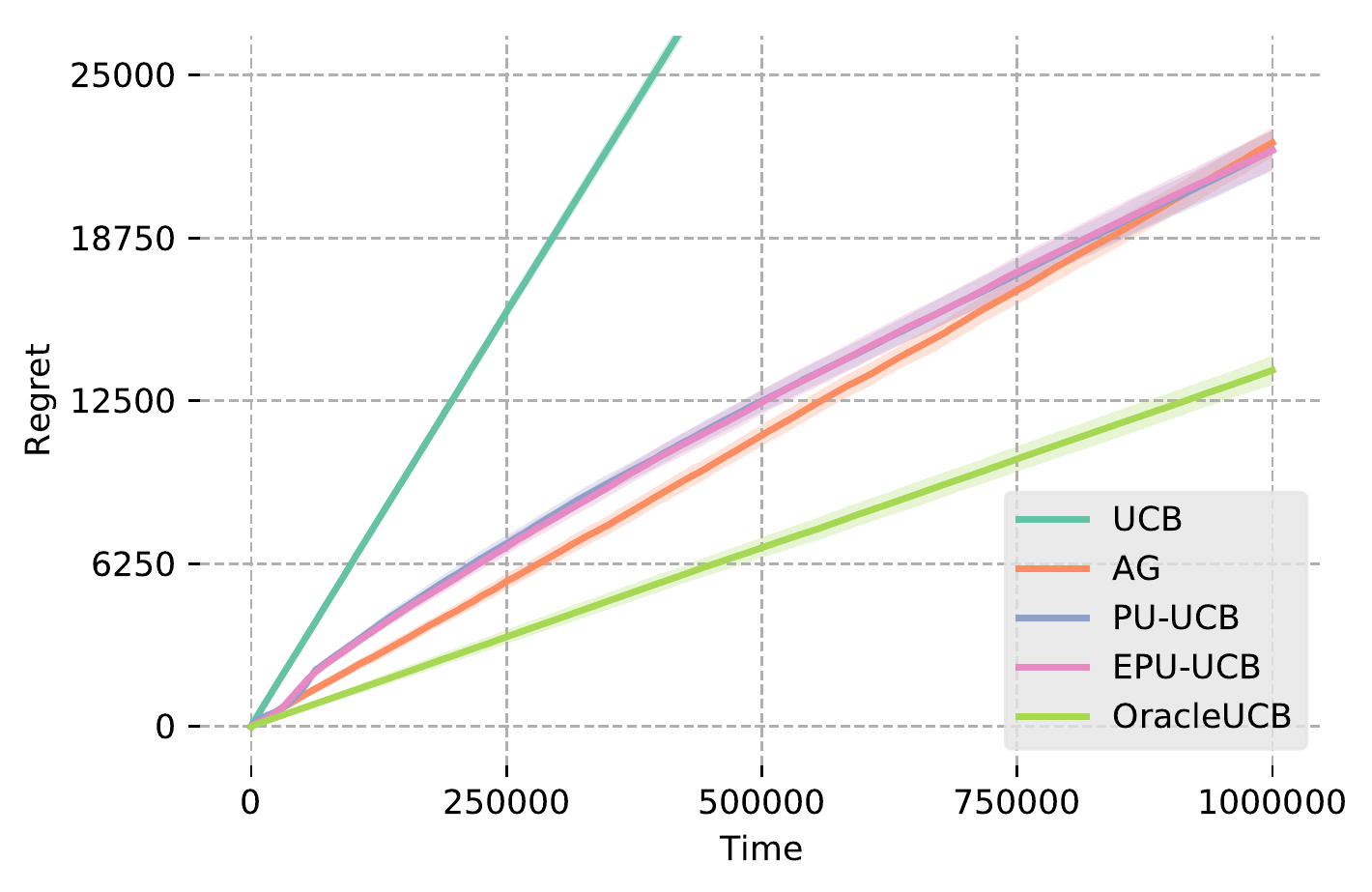}
    \caption{Uniform prior}
    \end{subfigure}
    \begin{subfigure}{0.49\textwidth}
    \includegraphics[width=0.99\linewidth]{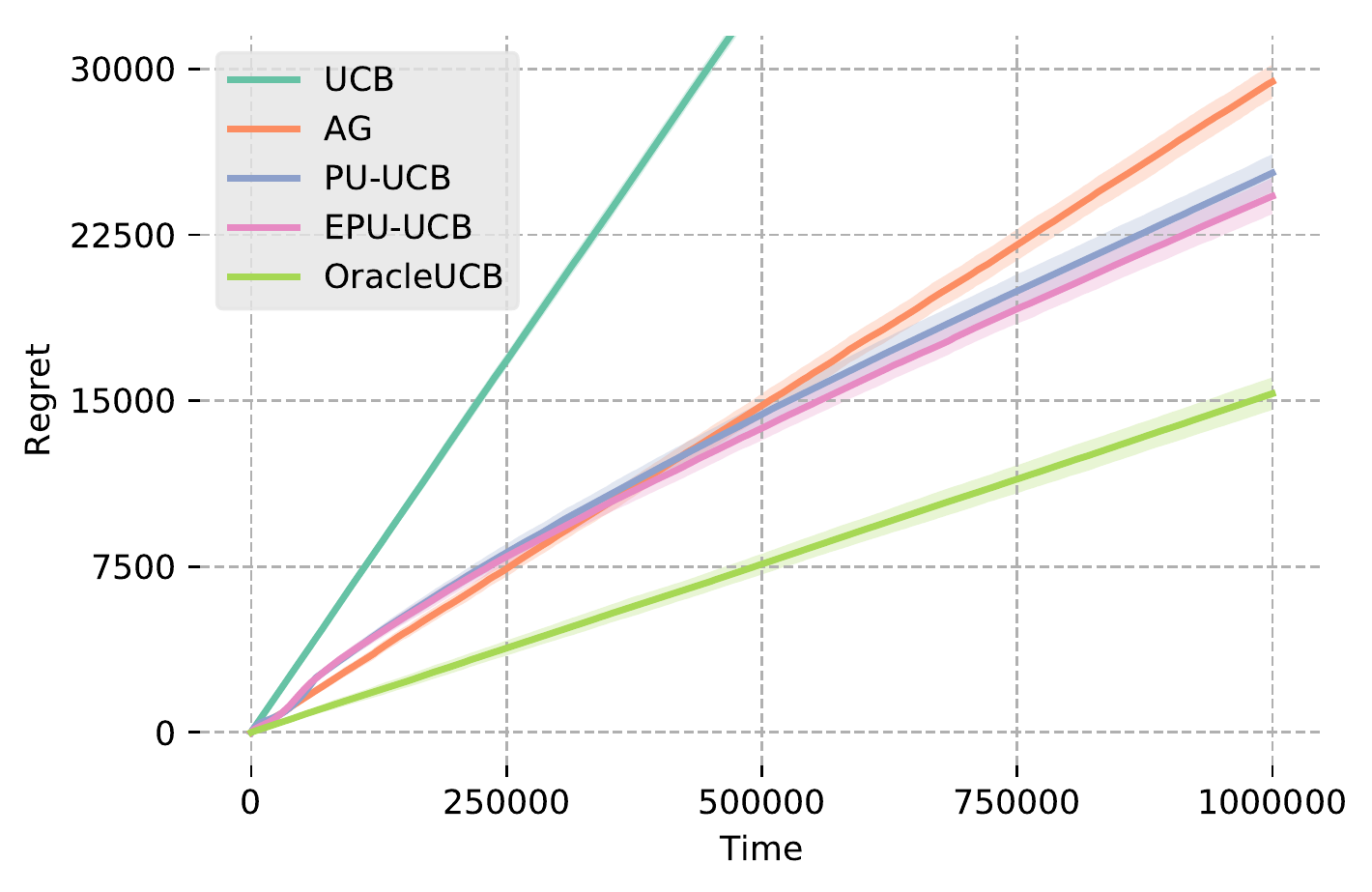}
    \caption{Beta(1, 3) prior} 
    \end{subfigure}
    \caption{Regret of various algorithms in the mortal bandit setting. Shaded areas show standard errors. \algo{PU-UCB} and \algo{PEU-UCB} are almost identical in (a).}
    \label{fig:mortal_bandits}
\end{figure}

We consider the same setting as \textcite{chakrabarti2009mortal}. Although they mostly consider a large number of arms, we previously illustrated in Section \ref{sec:choice_strategies} that this problem can be reduced via sub-sampling to a more tractable one, and with better performance as well. Therefore we focus our attention on problems with a small number of arms.
In this setting, the number of arms remains fixed throughout the time horizon $T$, that is when an arm dies, it is immediately replaced by another one. The lifetime of arm $k$, denoted $L_k$, is drawn i.i.d.\ from a Geometric distribution with mean lifetime $L$; this arm died after being available for $L_k$ rounds. We also assume that arms are Bernoulli random variables. We consider two scenarios: in the first one, mean rewards of arms are drawn i.i.d.\ from a uniform distribution over [0, 1], while in the second scenario they are drawn from a Beta(1, 3) distribution. In both cases, we fix the number of arms $K=5$ and the expected lifetime $L=1000$. The horizon is set at $T=1000L$, meaning that there are on average 1000 episodes throughout the time horizon. We compare several algorithms: the untuned \algo{UCB} algorithm and its optimally tuned variant \algo{OracleUCB}, along with a tuned \algo{AdaptiveGreedy} (\algo{AG}) \cite{chakrabarti2009mortal};\footnote{Both algorithms have been tuned with respect to the mean lifetime. On scenario 1, $\gamma=0.25$ and $c=1.5$, while on scenario 2, $\gamma=0.25$ and $c=2.5$.} and our proposed methods, \algo{PeriodicUpdate-UCB} (\algo{PU-UCB}) and \algo{EstimatedPeriodicUpdate-UCB} (\algo{EPU-UCB}), where in the first one we assume the knowledge of the expected lifetime, while in the second it is estimated by the empirical mean of dead arms. On top of both our proposed algorithms lie a \algo{Greedy} algorithm on a discretization of $n=\sqrt{1000}$ points. Results are averaged over 100 runs and are reported on Figure \ref{fig:mortal_bandits}.
 
These experiments further underline the benefit of tuning the \algo{UCB} algorithm for practical purposes. \algo{AG}, a state of the art algorithm in the mortal bandit setting, is clearly outperformed by \algo{OracleUCB}.
Additionally, \algo{PU-UCB} and \algo{EPU-UCB}, with somewhat similar performances on both setups, become rapidly better than \algo{AG}, despite that the latter being optimally tuned. We also remark once more that the regret of both algorithms is sublinear, highlighting the performance of the \algo{Greedy} meta-algorithm. Interestingly, \algo{EPU-UCB} performs slightly better than \algo{PU-UCB} on scenario 2, hinting at a potentially better decomposition of episodes.


\section{Conclusion}


In this paper, we studied a lifelong learning problem in the multi-armed bandit framework where tasks arrive sequentially, sampled from a problem class with some prior distribution. We first introduced our method which consists in optimize a bandit algorithm, focusing on confidence interval width tuning of \algo{UCB}-like algorithms. We then considered a bandit over bandit approach employing greedy algorithms as meta-algorithm and evaluated them in both stationary and non-stationary environments.
Finally, we applied our method to a more realistic setting, the mortal bandit problem, by decomposing the time horizon into episodes according to the mean lifetime of arms and showed great empirical improvement compared to previous work. 

\paragraph{Interesting directions}
The most prominent future work, especially in terms of practical applications, concerns the tracking of seasonal environments. It is crucial for building lifelong learning agents. Just recently this problem has been studied in the non-stationary bandit setting \cite{di2020linear, chen2020learning}.

Another direction may be the analysis of the \algo{Greedy} algorithm in the continuous-armed bandit problem. It has been shown in a recent line a work that it enjoys great performances in several bandit frameworks \cite{bastani2017mostly, kannan2018smoothed, bayati2020optimal, raghavan2020greedy}, and this setting is most likely one of them as shown indirectly in our experiments. Although it may be suboptimal, with high probability it concentrates quickly on arms with high expected rewards and thus works extremely well in practice. The next question is the ``optimal'' discretization for the greatest performance.

Finally in this first work, we focused solely on classical context-free bandits, but extending it to other settings that receive increasing interest, such as contextual  \cite{Rigollet}, combinatorial \cite{degenne1, perrault1} or multi-player bandits \cite{boursier1,boursier2}, seems also quite crucial and an interesting direction of research.




\begin{ack}
The research presented was supported by the French National Research Agency, under the project BOLD (ANR19-CE23-0026-04) and it was also supported in part by a public grant as part of the Investissement d'avenir project, reference ANR-11-LABX-0056-LMH, LabEx LMH, in a joint call with Gaspard Monge Program for optimization, operations research and their interactions with data sciences.
\end{ack}

\printbibliography

\newpage

\appendix
\section{Additional simulations}


In this section, we repeat the simulations of Sections \ref{sec:choice_strategies}, \ref{sec:fixed_init} and \ref{sec:fixed_learning} for different environments.

\subsection{Choice of the class of algorithms} \label{appendix:choice_class_strategies}

\begin{figure}[hbt]
    \begin{subfigure}{0.33\textwidth}
    \includegraphics[width=\linewidth]{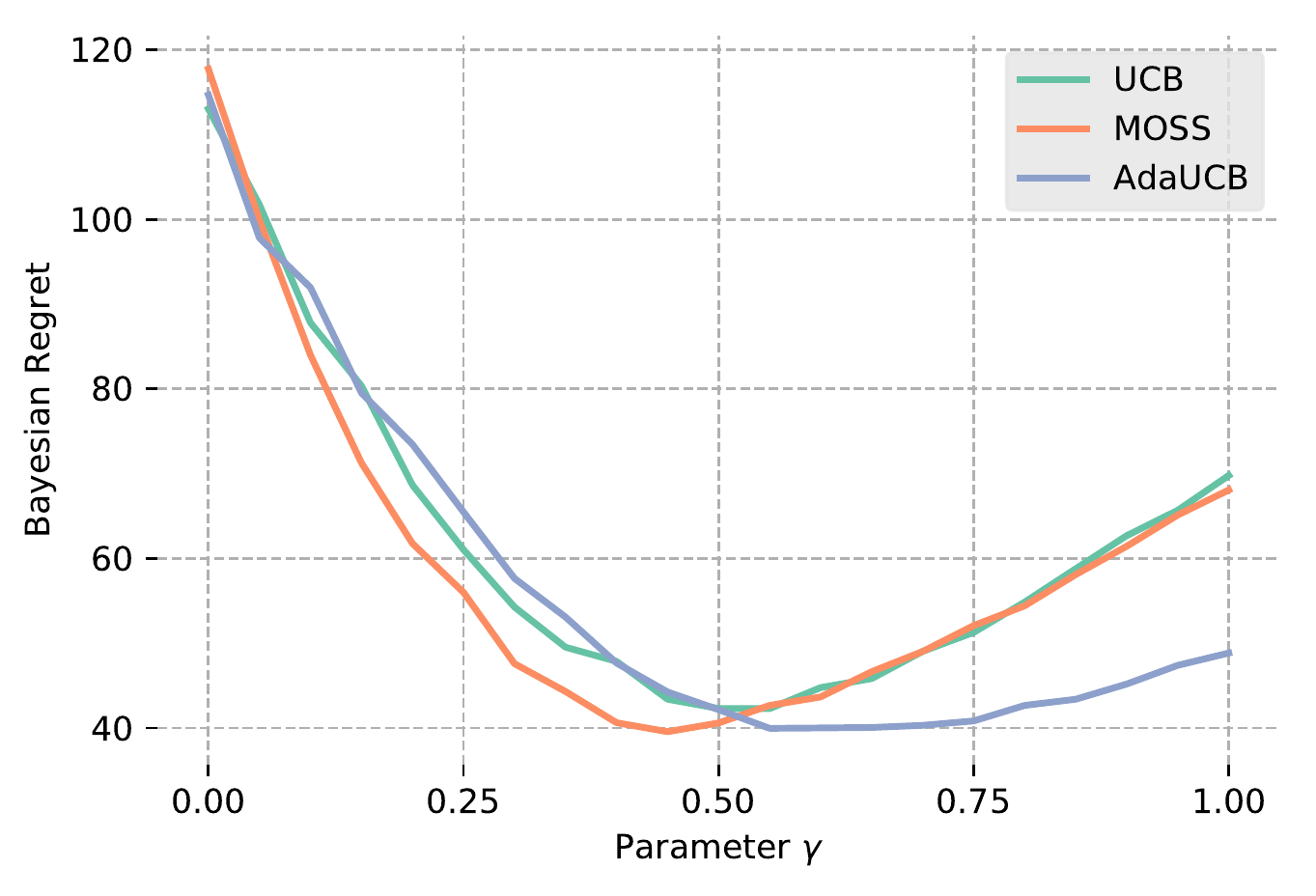}
    \caption{Scenario 1 / $K=5$}
    \end{subfigure}
    \begin{subfigure}{0.33\textwidth}
    \includegraphics[width=\linewidth]{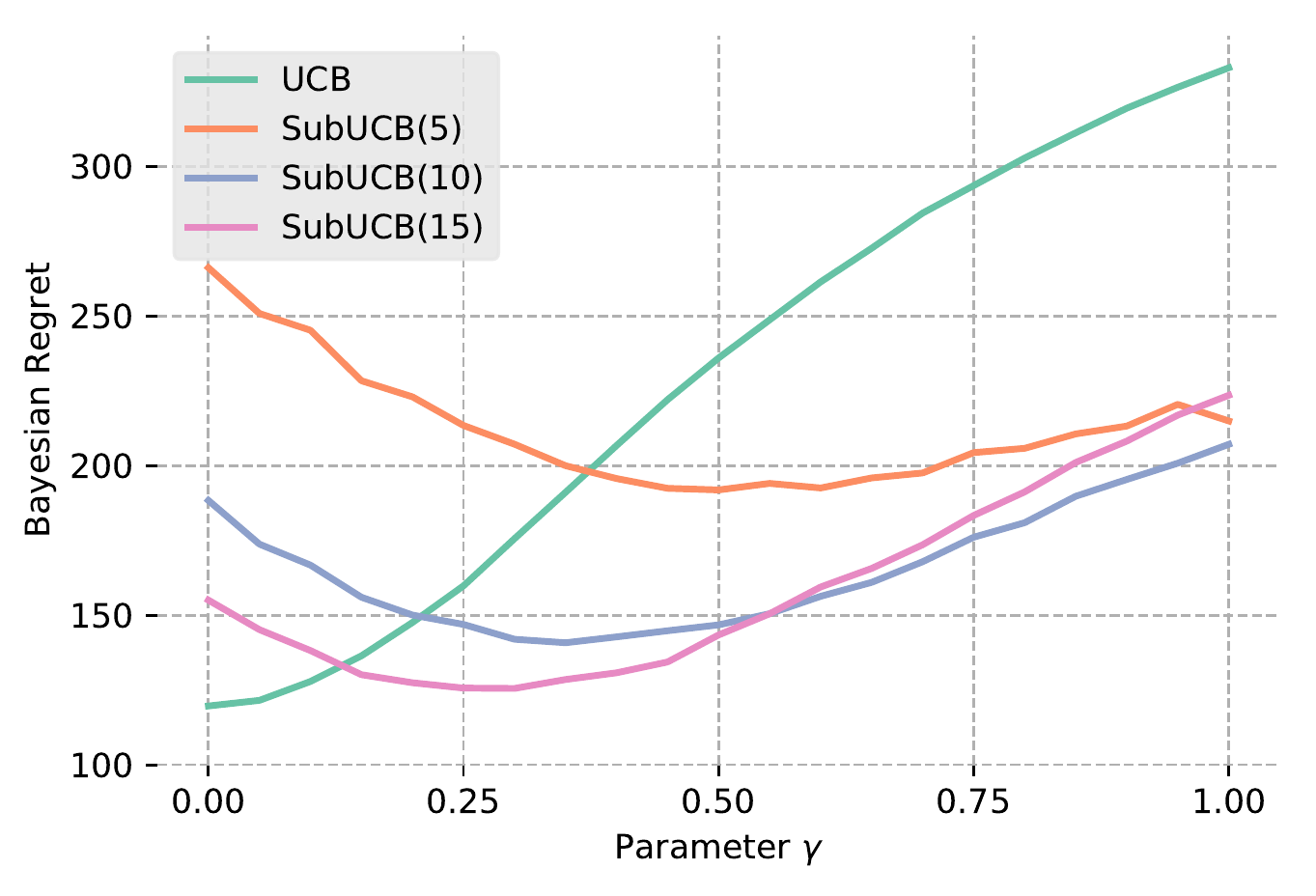}
    \caption{Scenario 1 / $K=63$}
    \end{subfigure}
    \begin{subfigure}{0.33\textwidth}
    \includegraphics[width=\linewidth]{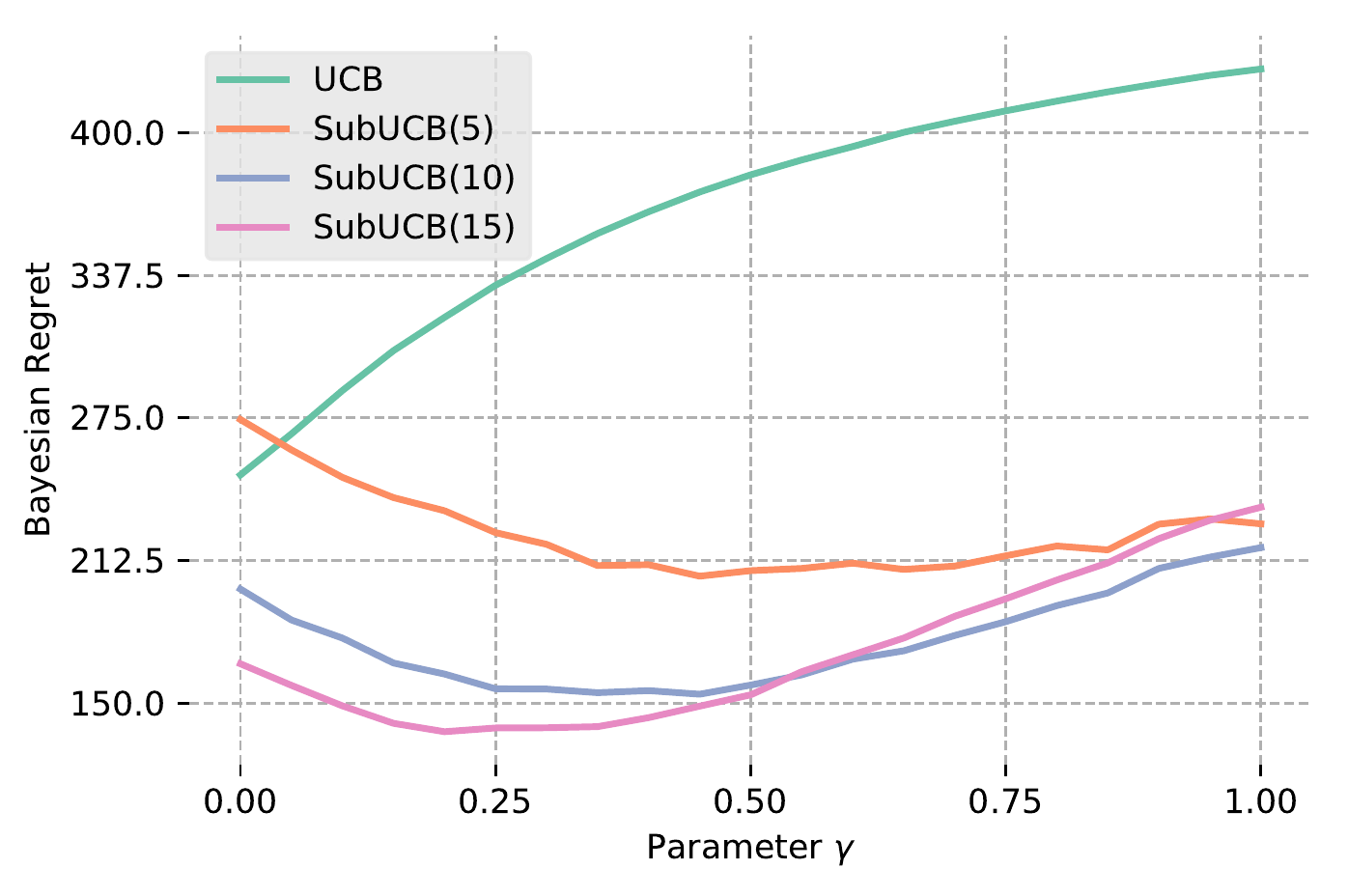}
    \caption{Scenario 1 / $K=250$}
    \end{subfigure}

    \begin{subfigure}{0.33\textwidth}
    \includegraphics[width=\linewidth]{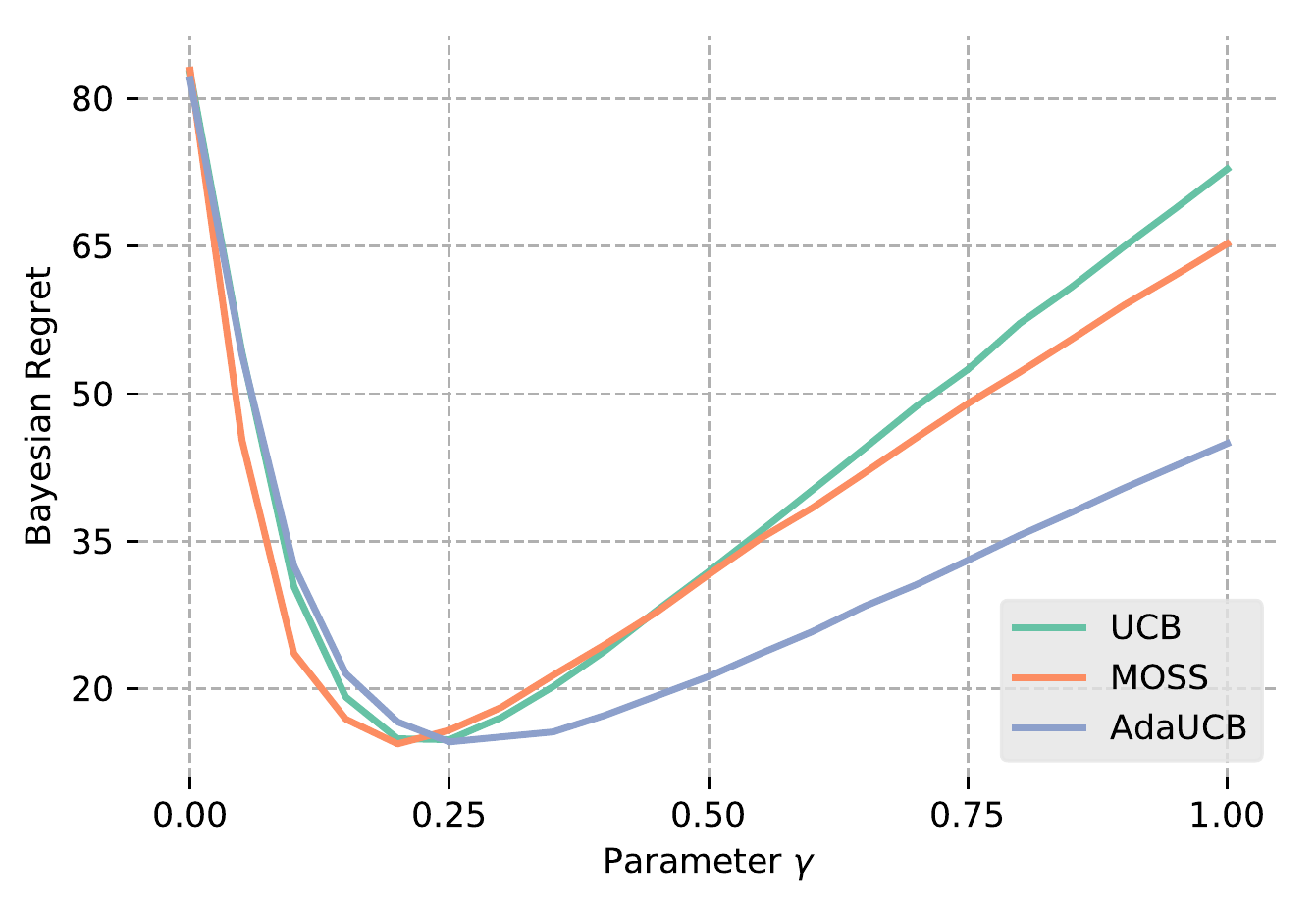}
    \caption{Scenario 2 / $K=5$}
    \end{subfigure}
    \begin{subfigure}{0.33\textwidth}
    \includegraphics[width=\linewidth]{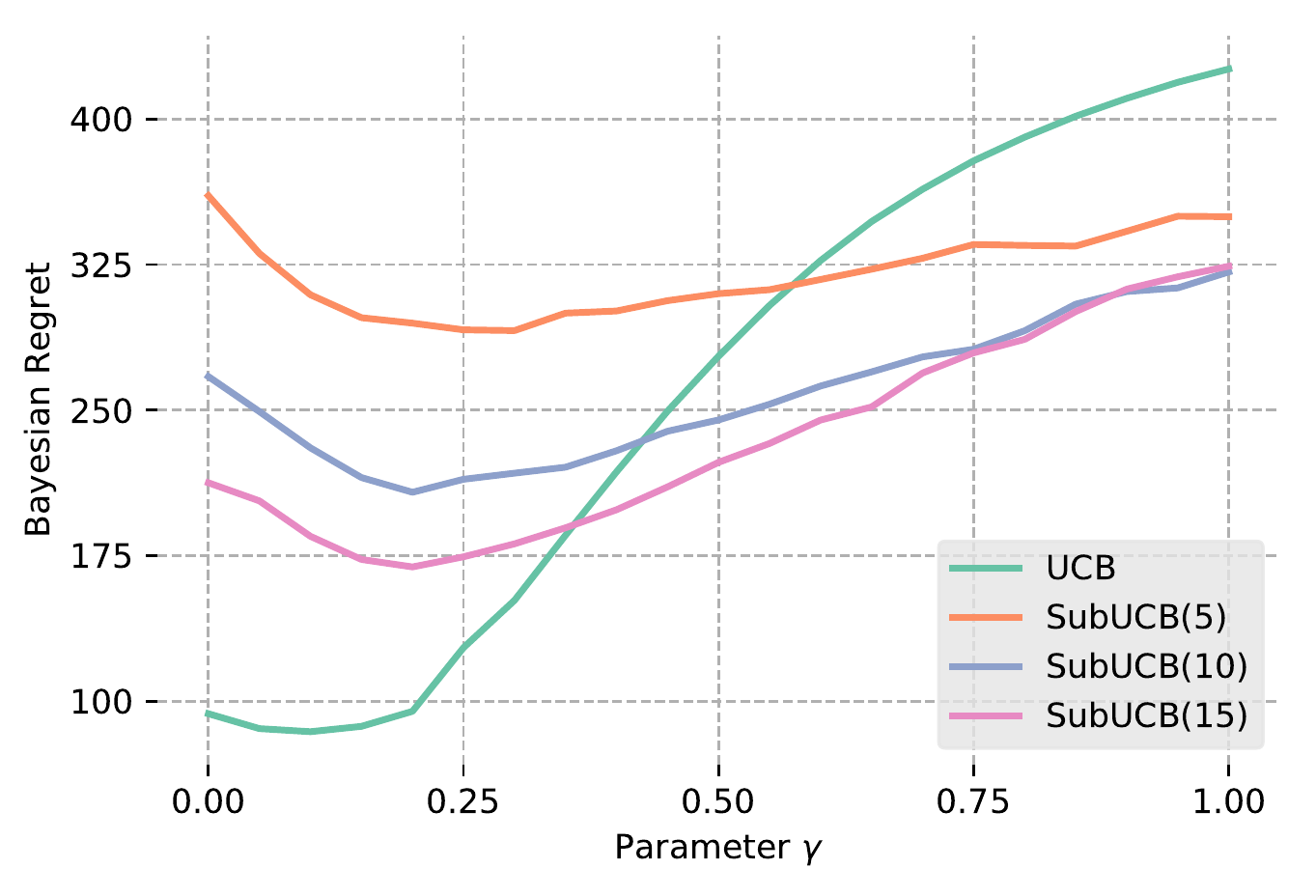}
    \caption{Scenario 2 / $K=63$}
    \end{subfigure}
    \begin{subfigure}{0.33\textwidth}
    \includegraphics[width=\linewidth]{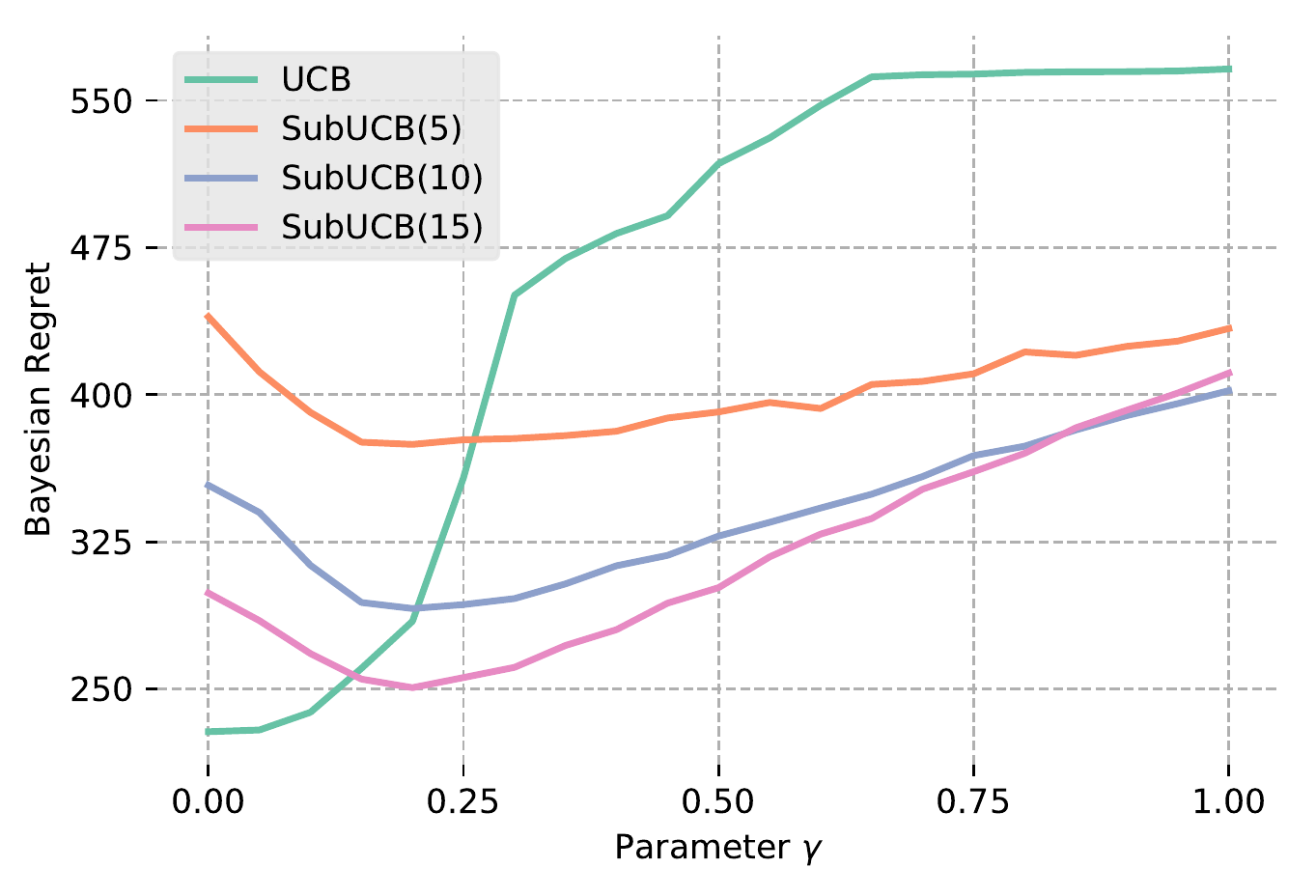}
    \caption{Scenario 2 / $K=250$}
    \end{subfigure}
    
    \caption{Bayesian regret of various algorithms as a function of $\gamma$ for diverse environments and numbers of arms $K$. Rows correspond respectively to Gaussian bandits with a uniform prior and Bernoulli bandits with a Beta(1, 3) prior.}
    \label{fig:choice_sub_policy_app}
\end{figure}

In this subsection, we repeat the simulations of Section \ref{sec:choice_strategies}. We recall that we evaluate the Bayesian regret of several \algo{UCB}-like algorithms as a function of their parameters $\gamma$. We consider two scenarios: the first scenario is a Gaussian bandit problem where mean rewards are drawn i.i.d.\ from a uniform distribution over $[0, 1]$; while the second scenario is a Bernoulli bandit problem where mean rewards are drawn i.i.d.\ from a Beta(1, 3) distribution. In all experiments, the horizon is fixed at $T= 1000$ and we vary the number of arms $K$. Results are averaged over $5000$ iterations and displayed on Figure \ref{fig:choice_sub_policy_app}.

In the first scenario, we observe a similar behavior compared to the Bernoulli case: for small numbers of arms $K$, \algo{AdaUCB} performs better than \algo{UCB} for all $\gamma$; for moderate values of $K$, the \algo{Greedy} algorithm is roughly the best; and for large values $K$, \algo{SubUCB} performs the best. Similarly, \algo{SubUCB}($m$) performs better as $m$ grows larger and becomes more sensitive to $\gamma$ at the same time.
In the second scenario, the same behavior can be noticed, however what we called moderate and large values of $K$ are, in this case, much higher than previously.

\subsection{Influence of the initialization} \label{appendix:init}

\begin{figure}[hbt]
    \begin{subfigure}{0.33\textwidth}
    \includegraphics[width=\linewidth]{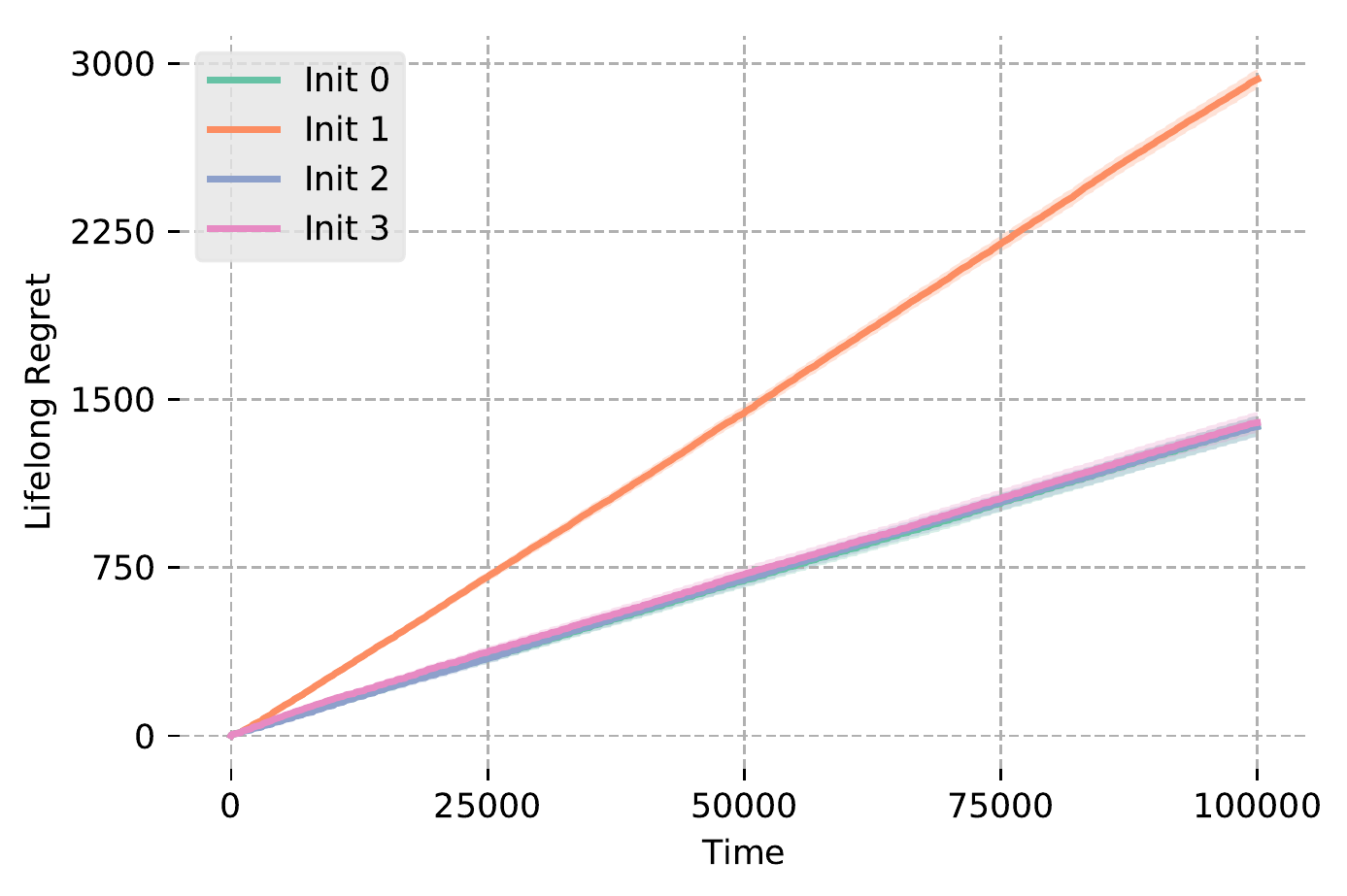}
    \caption{Scenario 1 / $K = 5$}
    \end{subfigure}
    \begin{subfigure}{0.33\textwidth}
    \includegraphics[width=\linewidth]{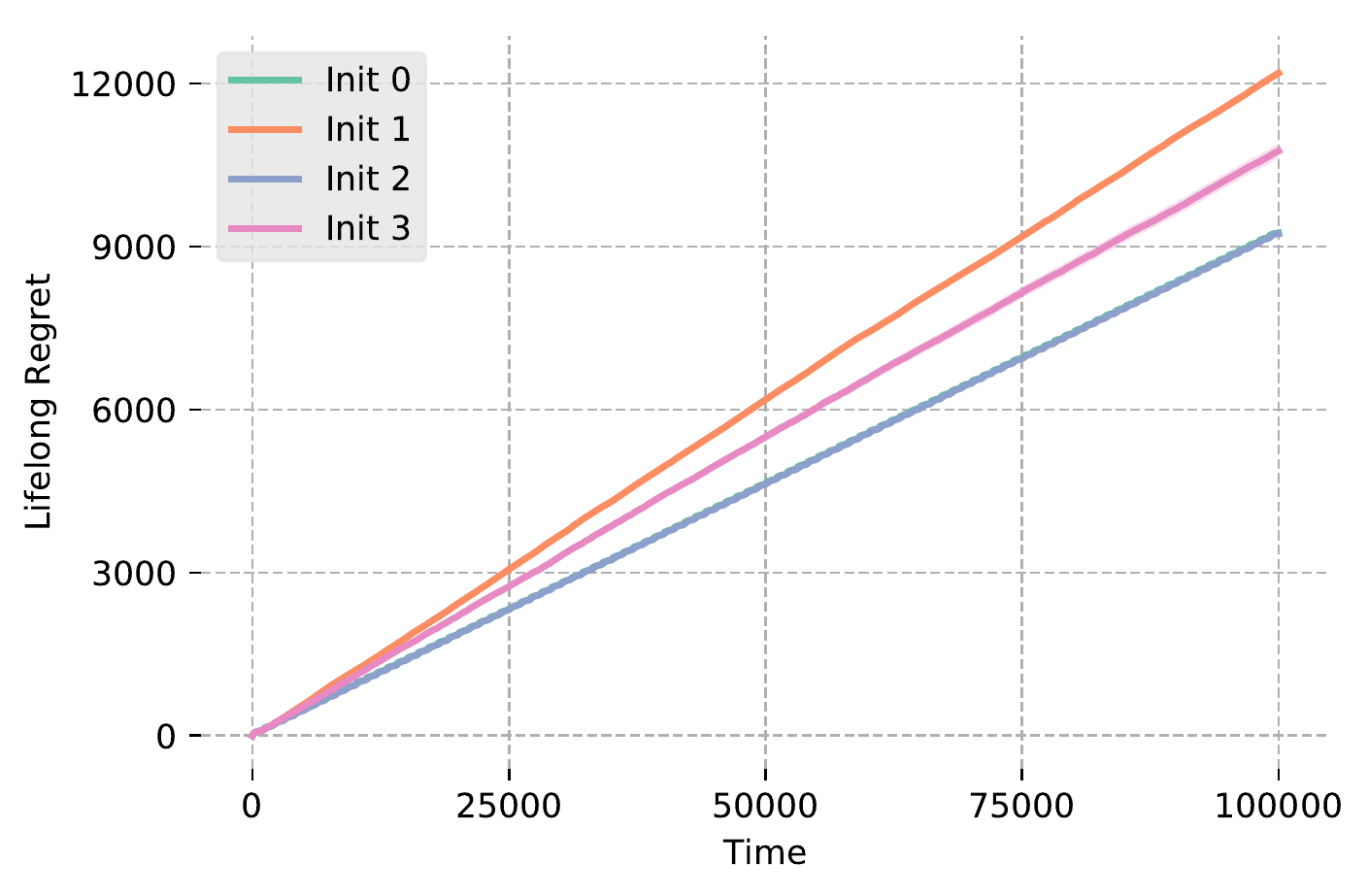}
    \caption{Scenario 1 / $K = 63$}
    \end{subfigure}
    \begin{subfigure}{0.33\textwidth}
    \includegraphics[width=\linewidth]{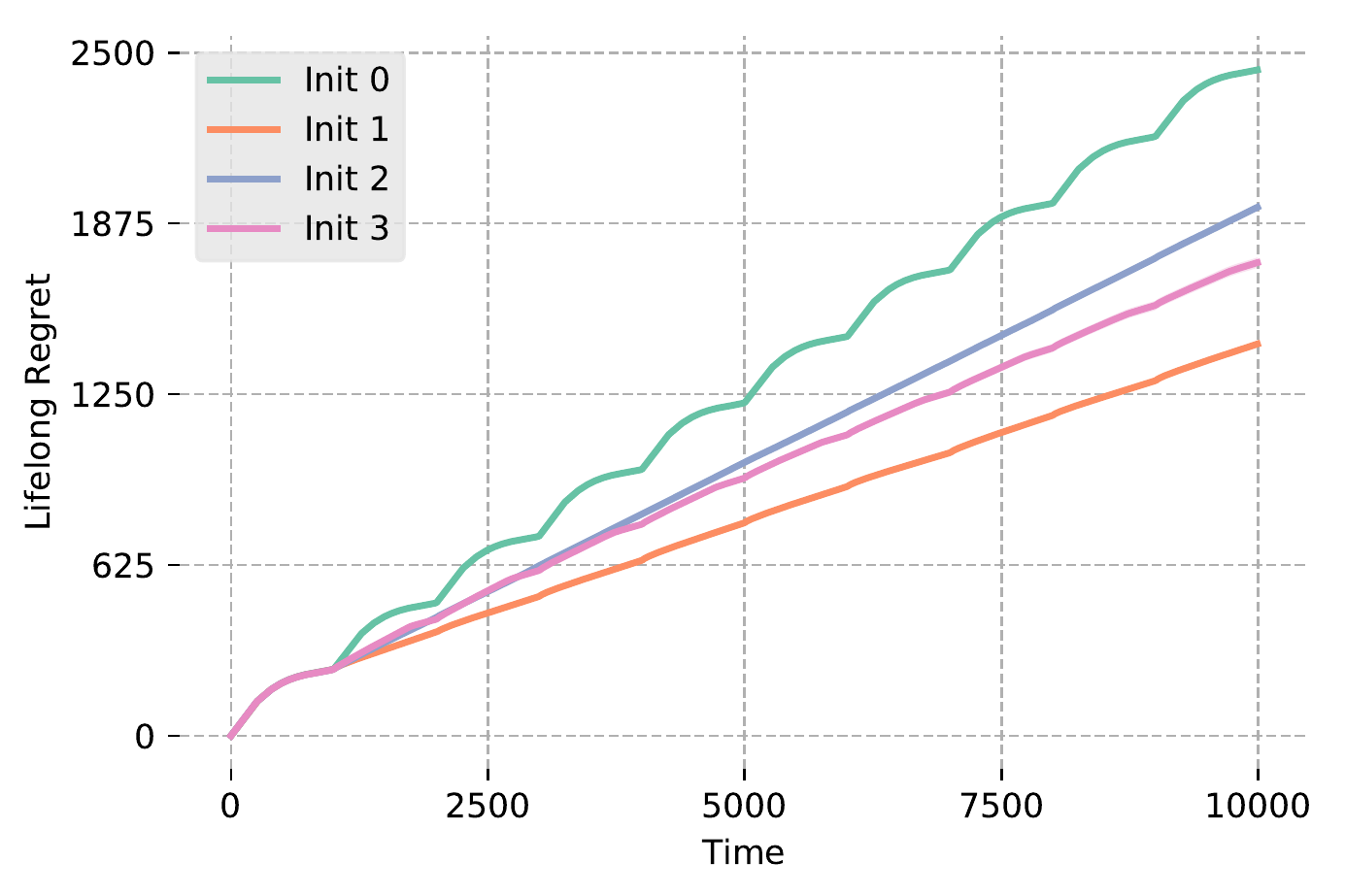}
    \caption{Scenario 1 / $K = 250$}
    \end{subfigure}
    
    \begin{subfigure}{0.33\textwidth}
    \includegraphics[width=\linewidth]{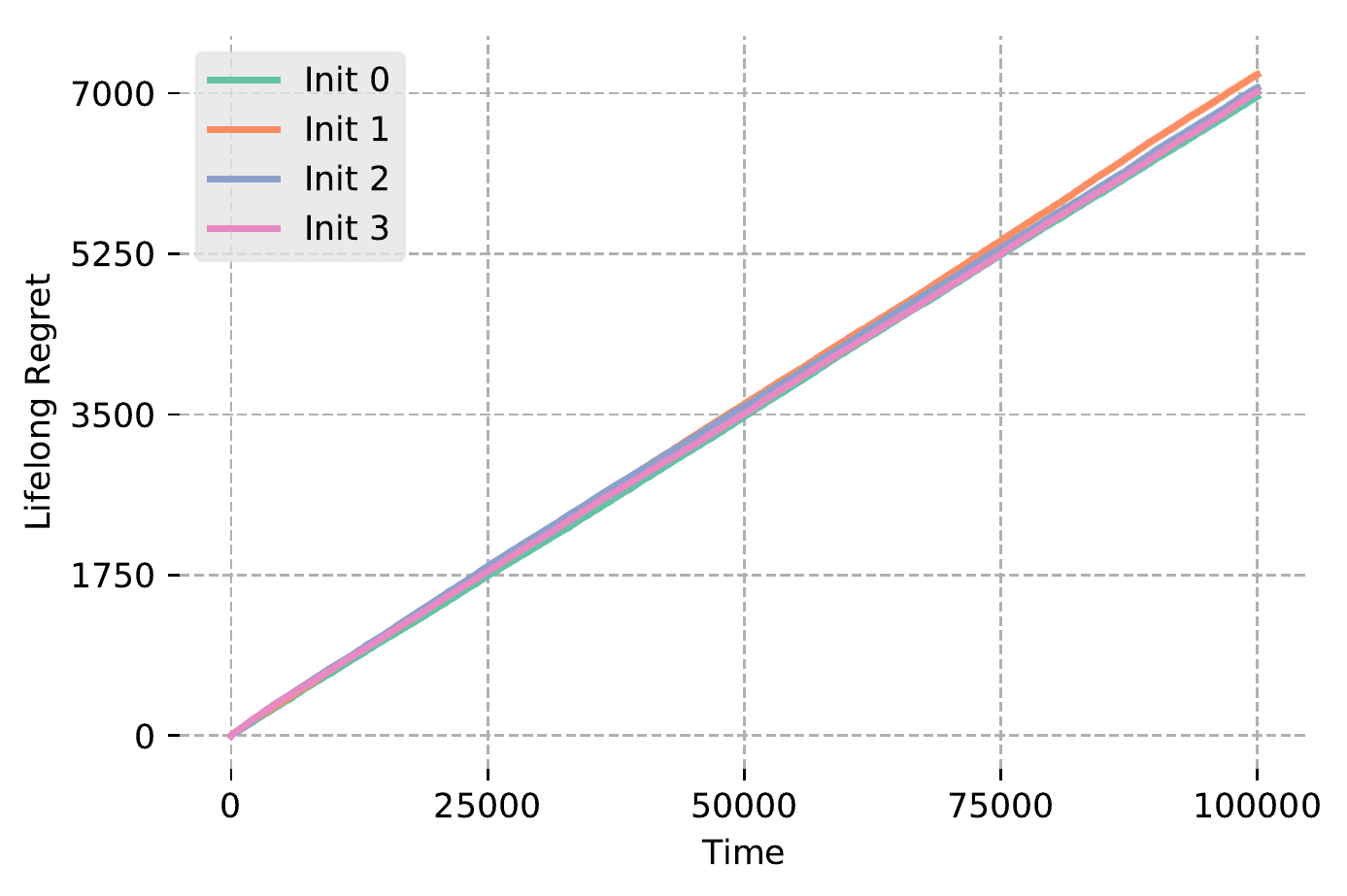}
    \caption{Scenario 2 / $K = 5$}
    \end{subfigure}
    \begin{subfigure}{0.33\textwidth}
    \includegraphics[width=\linewidth]{figures/init/scenario_1_K_63.pdf}
    \caption{Scenario 2 / $K = 63$}
    \end{subfigure}
    \begin{subfigure}{0.33\textwidth}
    \includegraphics[width=\linewidth]{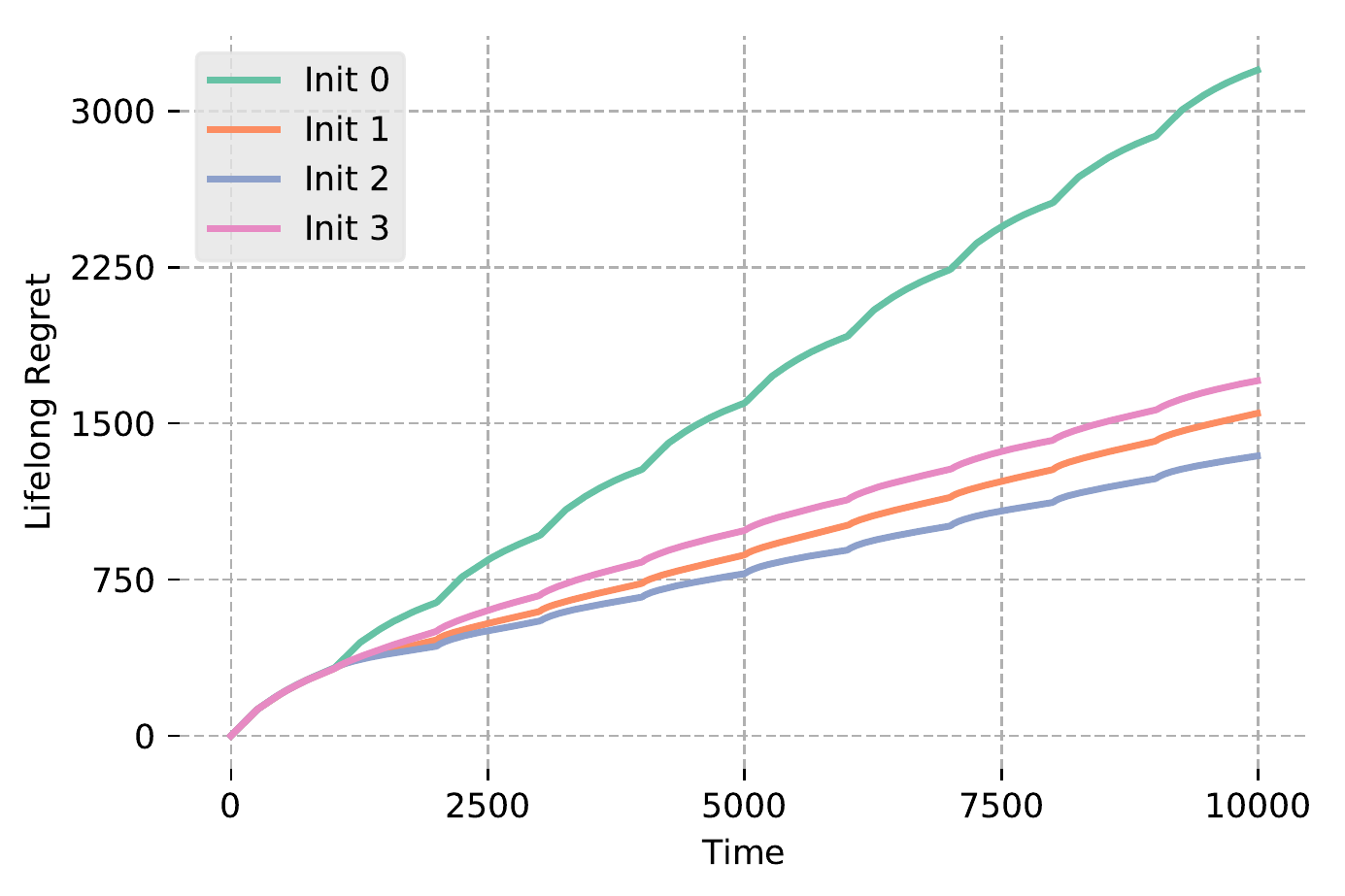}
    \caption{Scenario 2 / K = 250}
    \end{subfigure}
    
    \begin{subfigure}{0.33\textwidth}
    \includegraphics[width=\linewidth]{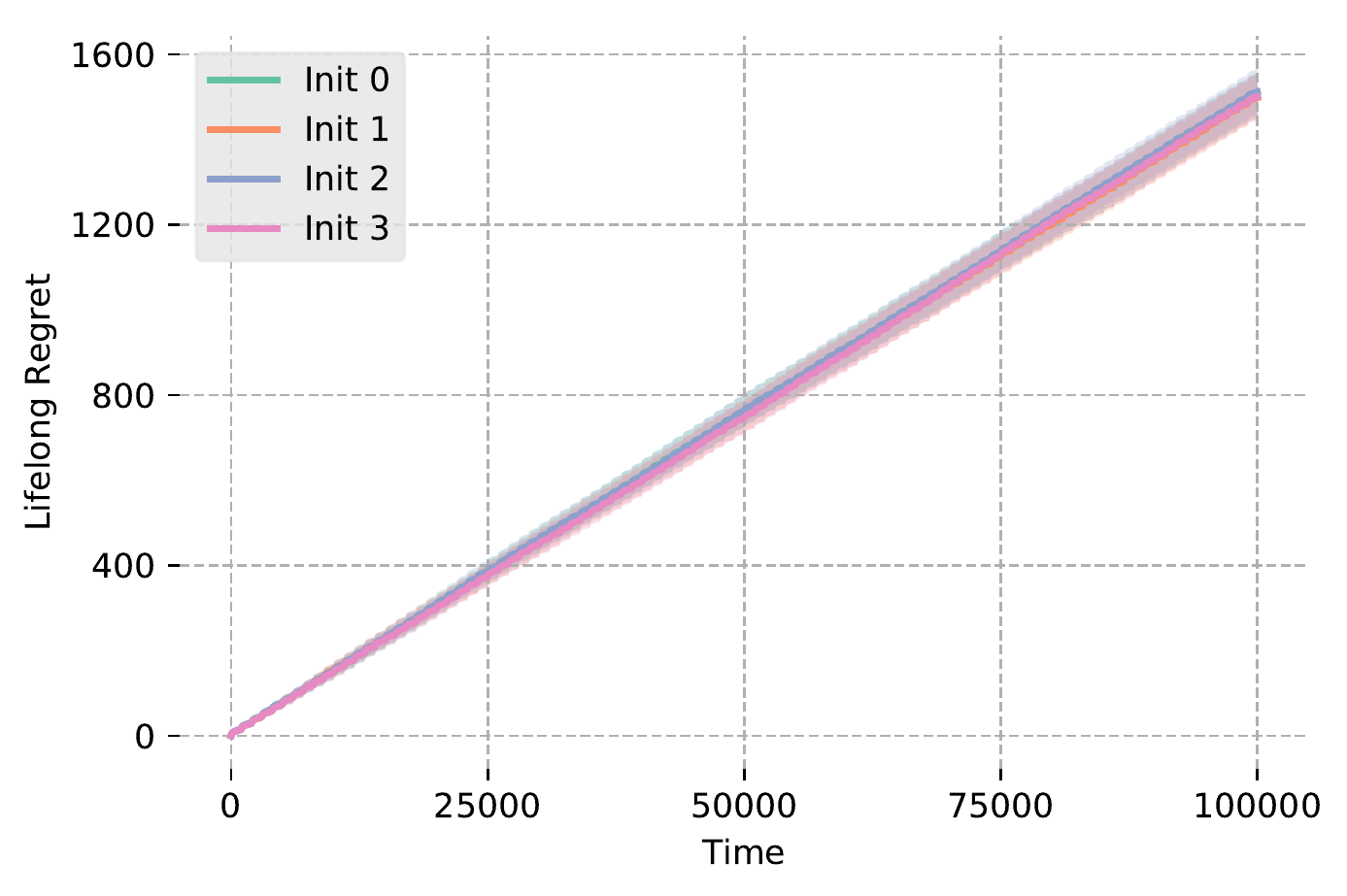}
    \caption{Scenario 3 / $K = 5$}
    \end{subfigure}
    \begin{subfigure}{0.33\textwidth}
    \includegraphics[width=\linewidth]{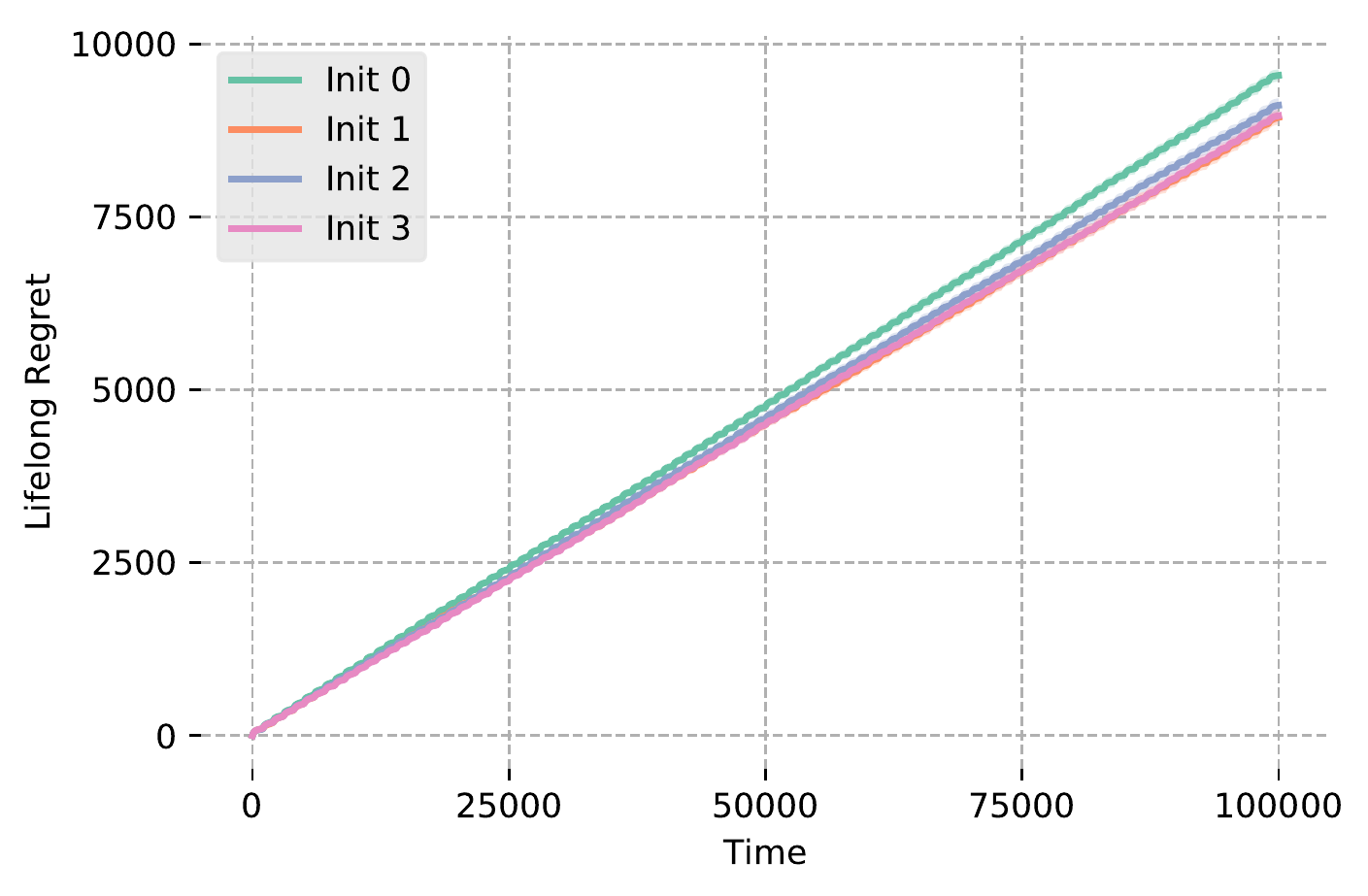}
    \caption{Scenario 3 / $K = 63$}
    \end{subfigure}
    \begin{subfigure}{0.33\textwidth}
    \includegraphics[width=\linewidth]{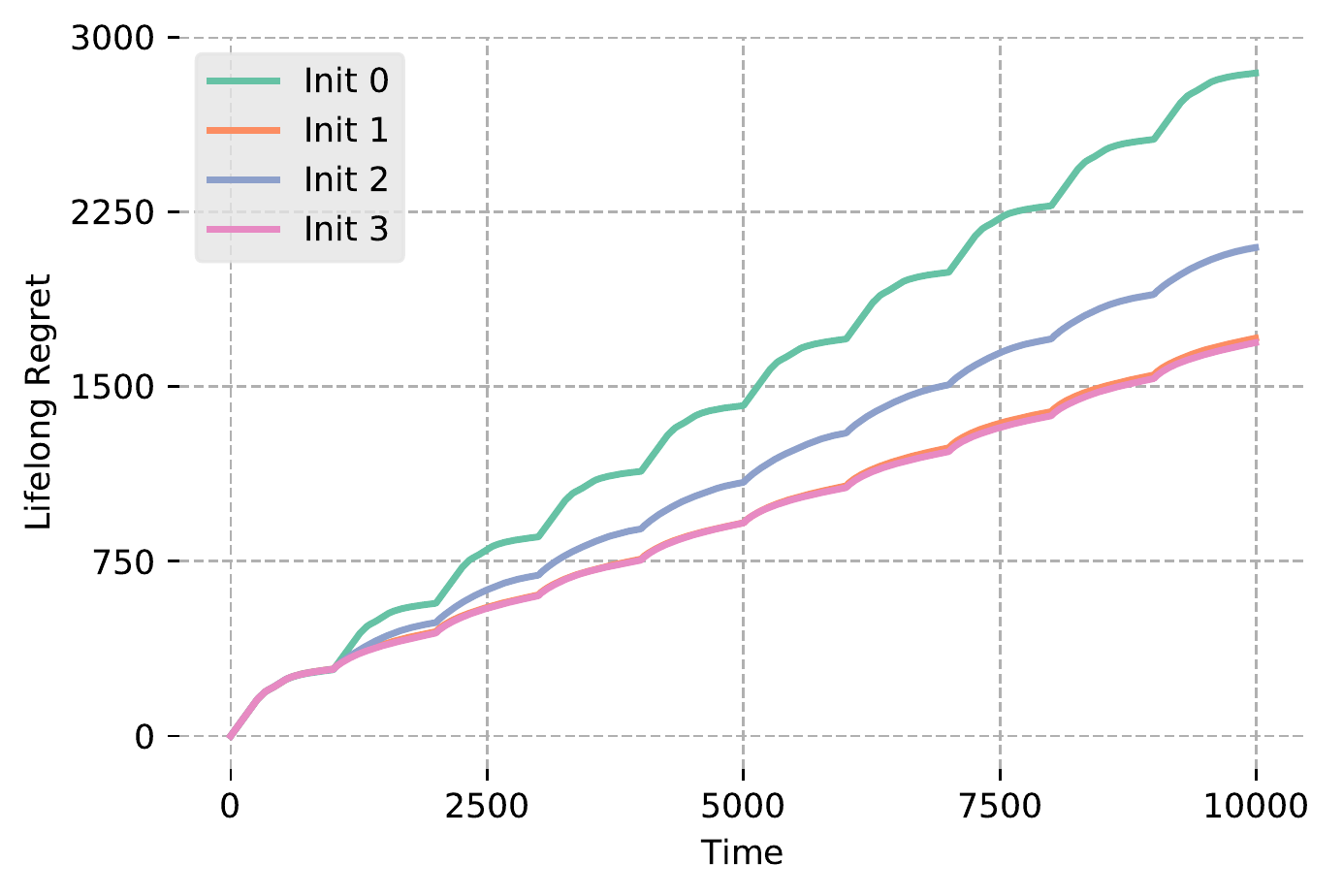}
    \caption{Scenario 3 / $K = 250$}
    \end{subfigure}
    \caption{Lifelong regret of a deterministic meta-algorithm with various initializations in stationary environments. Rows correspond respectively to Bernoulli bandits with a uniform prior, Gaussian bandits with a uniform prior and Bernoulli bandits with a Beta(1, 3) prior. Shaded areas show standard errors.}
    \label{fig:fixed_env_init_app}
\end{figure}

In this subsection, we repeat the simulations of Section \ref{sec:fixed_init}. 
We recall that we study the impact of several initializations on the lifelong regret. To do so, we set $\gamma = 0.22$ in the \algo{UCB} algorithm. We consider three scenarios: the first scenario is a Bernoulli bandit problem where mean rewards are drawn i.i.d.\ from a uniform distribution over $[0, 1]$, the second scenario is a Gaussian bandit problem where mean rewards are drawn i.i.d.\ from a uniform distribution over $[0, 1]$ and the third scenario is a Bernoulli bandit problem where mean rewards are drawn i.i.d.\ from a Beta(1, 3) distribution.
In all experiments, the horizon is fixed at $T= 1000$ and we vary the number of arms $K$. For $K=5$ and $K=63$, the numbers of episodes is $J=100$, whereas for $K=250$ it is $J=10$. 
Results are averaged over $500$ iterations for $K=5$, $100$ for the rest, and displayed on Figure \ref{fig:fixed_env_init_app}.

It is complex to observe a clear trend across the different simulations.
For small values of $K$, the choice of initialization is insignificant; except for the initialization at 0 in the first scenario, they all have roughly the same performance.
For intermediate values of $K$, the impact of the choice of initialization becomes apparent. Although there is no optimal choice, initializing arms with the median of previous arms seems more robust.
For large value of $K$, a clear trend is emerging: pulling each arm once is always the worst thing to do. It was expected since we spend most of the time initializing arms. There is still no optimal choice in this case, yet the initialization at 0 seems more robust.

\subsection{Learning in a stationary environment} \label{appendix:fixed_env}

\begin{figure}[hbt]
    \begin{subfigure}{0.49\textwidth}
    \includegraphics[width=\linewidth]{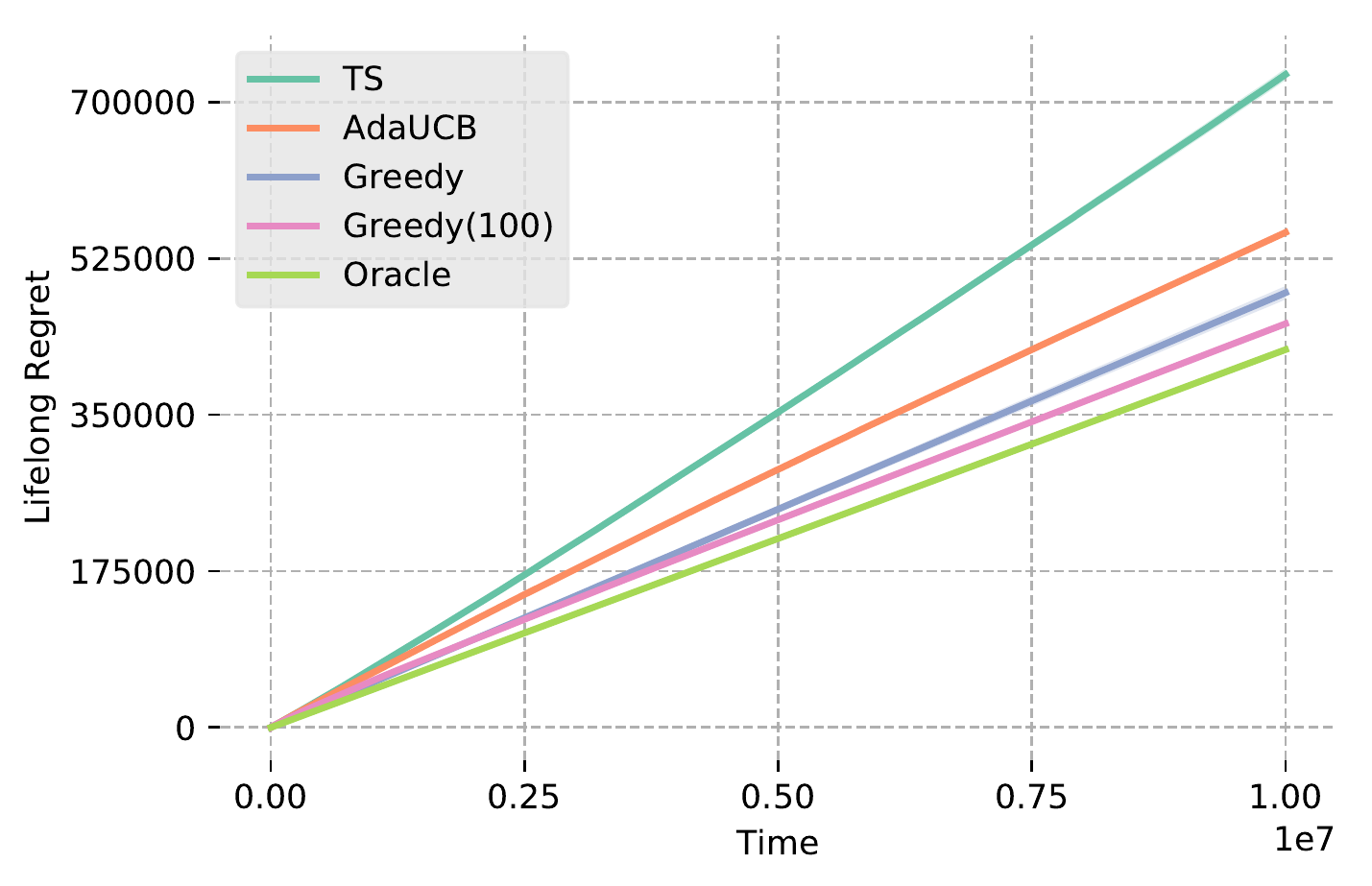}
    \caption{Gaussian bandits with uniform prior}
    \end{subfigure}
    \begin{subfigure}{0.49\textwidth}
    \includegraphics[width=\linewidth]{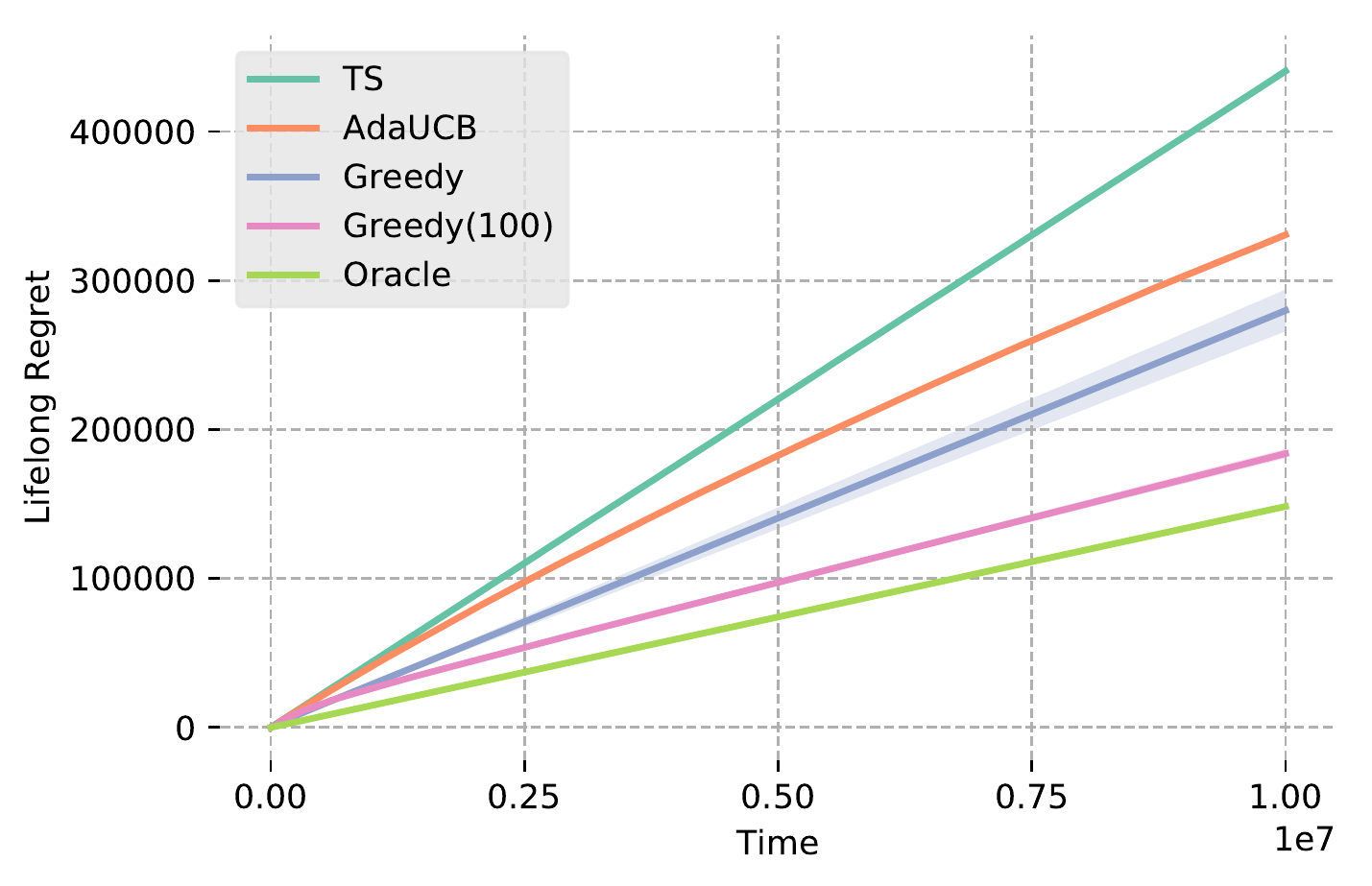}
    \caption{Bernoulli bandits with Beta(1, 3) prior}
    \end{subfigure}
    \caption{Lifelong regret of various meta-algorithms in stationary environments. Shaded areas show standard errors.}
    \label{fig:fixed_env_learning_app}
\end{figure}

In this subsection, we repeat the simulations of Section \ref{sec:fixed_learning}. We recall that we study the impact of several meta-algorithms on the lifelong regret. We consider two scenarios: the first scenario is a Gaussian bandit problem where mean rewards are drawn i.i.d.\ from a uniform distribution over $[0, 1]$; while the second scenario is a Bernoulli bandit problem where mean rewards are drawn i.i.d.\ from a Beta(1, 3) distribution. In all experiments and for each episode, the horizon is fixed at $T= 1000$ and the number of arms at $K=5$; and for the meta-algorithm, the number of episodes is set at $J=10000$. Results are averaged over $100$ iterations and illustrated on Figure \ref{fig:fixed_env_learning_app}.

The results are similar to Section \ref{sec:fixed_learning}: \algo{TS} fails to learn in that time frame, while \algo{AdaUCB} manages to do so but is outperformed for a long period of time by a naive \algo{Greedy}. As for \algo{Greedy}(100), it still performs extremely well; its regret is close to the optimal one and is again sublinear. 


\end{document}